\pdfoutput=1

\documentclass[journal]{IEEEtran}
\usepackage{cite}

\ifCLASSINFOpdf
\else
\fi

\usepackage{multicol}

\usepackage{graphicx} 
\usepackage{tabularx}
\usepackage{stfloats}
\usepackage{epstopdf}
\usepackage{booktabs}
\usepackage{array}
\newcolumntype{P}[1]{>{\centering\arraybackslash}p{#1}}
\usepackage{color}
\usepackage{colortbl}
\usepackage{multirow}
\usepackage{balance}
\usepackage{mathptmx} 
\usepackage{times} 
\usepackage{amsmath} 
\usepackage{amssymb}  
\usepackage{comment}
\usepackage{gensymb}
\usepackage{interval}
\usepackage{subcaption}
\usepackage{siunitx}
\usepackage{algorithm}
\usepackage{wrapfig}
\usepackage{pgfplots}
\pgfplotsset{compat=newest}
\usetikzlibrary{plotmarks}
\usetikzlibrary{babel}
\usetikzlibrary{arrows.meta}
\usepgfplotslibrary{patchplots}
\usepackage{grffile}
\usepackage{amsmath}

\usepackage{hyperref}

\definecolor{alexey}{rgb}{1.0,0.0,0.0}
\definecolor{rene}{rgb}{1.0,0.0,0.0}
\definecolor{davide}{rgb}{1.0,0.0,0.0}
\definecolor{cready}{rgb}{0.0,0.0,0.0}

\newcommand\cready[1]{\textcolor{cready}{#1}}

\definecolor{somegray}{rgb}{0.5, 0.5, 0.5}
\newcommand{\darkgrayed}[1]{\textcolor{somegray}{#1}}
\makeatletter
\newcommand*\titleheader[1]{\gdef\@titleheader{#1}}
\AtBeginDocument{%
  \let\st@red@title\@title
  \def\@title{%
    \vskip-2em
    \bgroup\normalfont\large\centering\@titleheader\par\egroup
    \vskip1.5em\st@red@title}
}
\makeatother

\titleheader{\darkgrayed{This paper has been accepted for publication in the\\
IEEE Transaction on Robotics (T-RO) journal, 2019.
\copyright IEEE}}

\title{Deep Drone Racing: from Simulation to Reality with Domain Randomization}

\author{\IEEEauthorblockN{Antonio Loquercio\IEEEauthorrefmark{1}\IEEEauthorrefmark{3},
        Elia Kaufmann\IEEEauthorrefmark{1}\IEEEauthorrefmark{3}, 
		Ren\'{e} Ranftl\IEEEauthorrefmark{2},
		Alexey Dosovitskiy\IEEEauthorrefmark{2},
		Vladlen Koltun\IEEEauthorrefmark{2}, and\\
        Davide Scaramuzza\IEEEauthorrefmark{1}}%
\IEEEauthorblockA{\thanks{\IEEEauthorrefmark{1}The authors are with the Robotic and Perception
Group, at both the Dep. of Informatics (University of Zurich) and the Dep.
of Neuroinformatics (University of Zurich and ETH Zurich), Andreasstrasse
15, 8050 Zurich, Switzerland.
}}
\IEEEauthorblockA{\thanks{\IEEEauthorrefmark{2}The authors are with the Intelligent Systems Lab, Intel.}
}
\IEEEauthorblockA{\thanks{\IEEEauthorrefmark{3}These authors contributed equally.}
}}

\begin{document}

\maketitle

\begin{abstract}
Dynamically changing environments, unreliable state estimation, and operation under severe resource constraints are fundamental challenges that limit the deployment of small autonomous drones.
We address these challenges in the context of autonomous, vision-based drone racing in dynamic environments.
A racing drone must traverse a track with possibly moving gates at high speed.
We enable this functionality by combining the performance of a state-of-the-art planning and control system with the perceptual awareness of a convolutional neural network (CNN).
The resulting modular system is both platform- and domain-independent: it is trained in simulation and deployed on a physical quadrotor without any fine-tuning.
The abundance of simulated data, generated via domain randomization, makes our system robust to changes of illumination and gate appearance.
To the best of our knowledge, our approach is the first to demonstrate \emph{zero-shot sim-to-real} transfer on the task of agile drone flight.
We extensively test the precision and robustness of our system, both in simulation and on a physical platform, and show significant improvements over the state of the art. 
\end{abstract}


\IEEEpeerreviewmaketitle

\section*{Source code, videos, and trained models}
Supplementary videos, source code, and trained networks can be found on the project page: \url{http://rpg.ifi.uzh.ch/research\_drone\_racing.html}


\section{Introduction}

\begin{figure}[t]
\centering
  \begin{subfigure}[b]{0.98\linewidth}
   \includegraphics[width=1\linewidth]{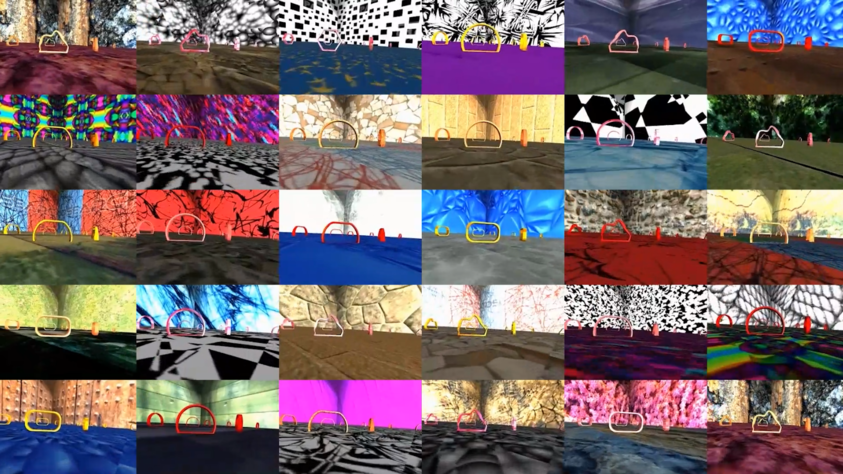}
   \vspace{1ex}
\end{subfigure}

\begin{subfigure}[b]{0.98\linewidth}
   \includegraphics[width=1\linewidth]{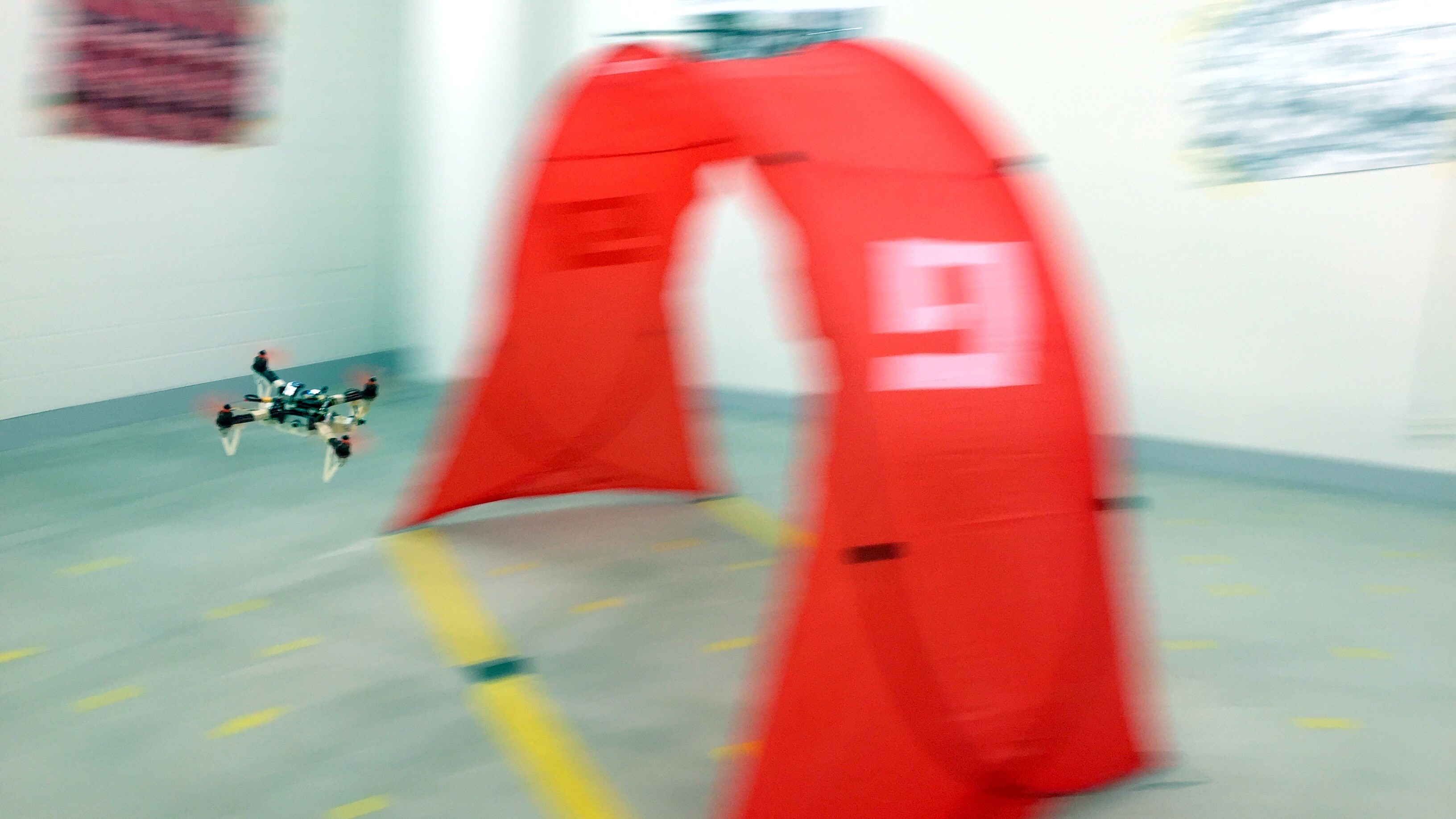}
\end{subfigure}

  \caption{\cready{The perception block of our system, represented by a convolutional neural network (CNN), is trained \emph{only} with non-photorealistic simulation data. 
  Due to the abundance of such data, generated with domain randomization, the trained CNN can be deployed on a physical quadrotor without any finetuning. }}
  
  \label{fig:catch_race}
\vspace{-5mm}
\end{figure}

\IEEEPARstart{D}{rone} racing is a popular sport in which professional pilots fly small quadrotors through complex tracks at high speeds (Fig.~\ref{fig:catch_race}).
Drone pilots undergo years of training to master the sensorimotor skills involved in racing.
Such skills would also be valuable to autonomous systems in applications such as disaster response or structure inspection, where drones must be able to quickly and safely fly through complex dynamic environments~\cite{yang2018grand}.

Developing a fully autonomous racing drone is difficult due to challenges that span dynamics modeling, onboard perception, localization and mapping, trajectory generation, and optimal control.
For this reason, autonomous drone racing has attracted significant interest from the research community, giving rise to multiple autonomous drone racing competitions~\cite{MooSunBalKim17, Moon2019}.

One approach to autonomous racing is to fly through the course by tracking a precomputed global trajectory.
However, global trajectory tracking requires to know the race-track layout in advance, along with highly accurate state estimation, which current methods are still not able to provide~\cite{svo, qin2018vins, Cadena2016}.
Indeed, visual inertial odometry~\cite{svo,qin2018vins} is subject to drift in estimation over time.
%
SLAM methods can reduce drift by relocalizing in a previously-generated, globally-consistent map.
However, enforcing global consistency leads to increased computational demands that strain the limits of on-board processing.
In addition, regardless of drift, both odometry and SLAM pipelines enable navigation only in a predominantly-static world, where waypoints and collision-free trajectories can be statically defined.
Generating and tracking a global trajectory would therefore fail in applications where the path to be followed cannot be defined \emph{a priori}. This is usually the case for professional drone competitions, since gates can be moved from one lap to another.

In this paper, we take a step towards autonomous, vision-based drone racing in dynamic environments.
Instead of relying on globally consistent state estimates, our approach deploys a convolutional neural network to identify waypoints in local body-frame coordinates.
This eliminates the problem of drift and simultaneously enables our system to navigate through dynamic environments.
The network-predicted waypoints are then fed to a state-of-the-art planner~\cite{mueller2013computationally} and tracker~\cite{faessler2018differential}, which generate a short trajectory segment and corresponding motor commands to reach the desired location.
The resulting system combines the perceptual awareness of CNNs with the precision offered by state-of-the-art planners and controllers, getting the best of both worlds.
The approach is both powerful and lightweight: all computations run fully onboard.

An earlier version of this work~\cite{kaufmann2018deep} (Best System Paper award at the Conference on Robotic Learning, 2018) demonstrated the potential of our approach both in simulation and on a physical platform.
In both domains, our system could perform complex navigation tasks, such as seeking a moving gate or racing through a dynamic track, with higher performance than state-of-the-art, highly engineered systems.
In the present paper, we extend the approach to generalize to environments and conditions not seen at training time.
In addition, we evaluate the effect of design parameters on closed-loop control performance, and analyze the computation-accuracy trade-offs in the system design.

In the earlier version~\cite{kaufmann2018deep}, the perception system was track specific: it required a substantial amount of training data from the target race track.
Therefore, significant changes in the track layout, background appearance, or lighting would hurt performance.
In order to increase the generalization abilities and robustness of our perception system, we propose to use domain randomization~\cite{tobin2017domain}.
The idea is to randomize during data collection all the factors to which the system must be invariant, i.e., \ illumination, viewpoint, gate appearance, and background.
We show that domain randomization leads to an increase in closed-loop performance relative to our earlier work~\cite{kaufmann2018deep} when evaluated in environments or conditions not seen at training time.
\cready{ Specifically, we demonstrate performance increases of up to $300\%$  in simulation (Fig.~\ref{fig:simulation_generalization}) and up to $36\%$ in real-world experiments (Fig.~\ref{fig:sim_vs_real}).}

Interestingly, the perception system becomes invariant not only to specific environments and conditions but also to the training domain.
We show that after training purely in non-photorealistic simulation, the perception system can be deployed on a physical quadrotor that successfully races in the real world.
On real tracks, the policy learned in simulation has comparable performance to one trained with real data, thus alleviating the need for tedious data collection in the physical world.

\section{Related Work}
\label{sec:rel_work}

Pushing a robotic platform to the limits of handling gives rise to fundamental challenges for both perception and control.
On the perception side, motion blur, challenging lighting conditions, and aliasing can cause severe drift in vision-based state estimation~\cite{svo, orb_slam, get_out_lab}.
Other sensory modalities, e.g.\ LIDAR or event-based cameras, could partially alleviate these problems~\cite{lidar_slam, Rosinol18ral}.
Those sensors are however either too bulky or too expensive to be used on small racing quadrotors.
Moreover, state-of-the-art state estimation methods are designed for a predominantly-static world, where no dynamic changes to the environment occur.

From the control perspective, plenty of work has been done to enable high-speed navigation, both in the context of autonomous drones~\cite{mellinger2011minimum, mueller2013computationally, morrell2018differential} and autonomous cars~\cite{kritayakirana2012autonomous, kapania2016trajectory, kegelman2017insights, WilDreGolRehThe16}.
However, the inherent difficulties of state estimation make these methods difficult to adapt for small, agile quadrotors that must rely solely on onboard sensing and computing.
We will now discuss approaches that have been proposed to overcome the aforementioned problems.

\subsection{Data-driven Algorithms for Autonomous Navigation}

A recent line of work, focused mainly on autonomous driving, has explored data-driven approaches that tightly couple perception and control~\cite{deep_drive, pan2018agile, safe_deep_nav, kahn2018self}.
These methods provide several interesting advantages, e.g.\ robustness against drifts in state estimation~\cite{deep_drive, pan2018agile} and the possibility to learn from failures~\cite{kahn2018self}.
The idea of learning a navigation policy end-to-end from data has also been applied in the context of autonomous, vision-based drone flight~\cite{cad2rl, dronet, gandhi2017learning}.
To overcome the problem of acquiring a large amount of annotated data to train a policy, Loquercio et al.~\cite{dronet}
proposed to use data from ground vehicles, while Gandhi et al.~\cite{gandhi2017learning} devised a method for automated data collection from the platform itself.
Despite their advantages, end-to-end navigation policies suffer from high sample complexity and low generalization to conditions not seen at training time.
This hinders their application to contexts where the platform is required to fly at high speed in dynamic environments.
To alleviate some of these problems while retaining the advantages of data-driven methods, a number of works propose to structure the navigation system into two modules: perception and control~\cite{hadsell2009learning, devin2017learning, chen2015deepdriving, Held-2017-102823, muller2018driving}.
This kind of modularity has proven to be particularly important for transferring sensorimotor systems across different tasks~\cite{Held-2017-102823, devin2017learning} and application domains~\cite{chen2015deepdriving, muller2018driving}.

We employ a variant of this perception-control modularization in our work.
However, in contrast to prior work, we enable high-speed, agile flight by making the output of our neural perception module compatible with fast and accurate model-based trajectory planners and trackers.

\subsection{Drone Racing}

The popularity of drone racing has recently kindled significant interest in the robotics research community.
The classic solution to this problem is image-based visual servoing, where a robot is given a set of target locations in the form of reference images or patterns. Target locations are then identified and tracked with hand-crafted detectors~\cite{classic_servo, aggressive_falanga, li2018autonomous}.
However, the handcrafted detectors used by these approaches quickly become unreliable in the presence of occlusions, partial visibility, and motion blur.
To overcome the shortcomings of classic image-based visual servoing, recent work proposed to use a learning-based approach for localizing the next target~\cite{jung2018perception}.
The main problem of this kind of approach is, however, limited agility.
Image-based visual servoing is reliable when the difference between the current and reference images is small, which is not always the case under fast motion.

Another approach to autonomous drone racing is to learn end-to-end navigation policies via imitation learning~\cite{mueller2017teaching}.
Methods of this type usually predict low-level control commands, in the form of body-rates and thrust, directly from images.
Therefore, they are agnostic to drift in state estimation and can potentially operate in dynamic environments, if enough training data is available.
However, despite showing promising results in simulated environments, these approaches still suffer from the typical problems of end-to-end navigation: (i) limited generalization to new environments and platforms and (ii) difficulties in deployment to real platforms due to high computational requirements (desired inference rate for agile quadrotor control is much higher than what current on-board hardware allows).

To facilitate robustness in the face of unreliable state estimation and dynamic environments, while also addressing the generalization and feasibility challenges, we use modularization. On one hand, we take advantage of the perceptual awareness of CNNs to produce navigation commands from images. On the other hand, we benefit from the high speed and reliability of classic control pipelines for generation of low-level controls.

\subsection{Transfer from Simulation to Reality}

Learning navigation policies from real data has a shortcoming: high cost of generating training data in the physical world.
Data needs to be carefully collected and annotated, which can involve significant time and resources.
To address this problem, a recent line of work has investigated the possibility of training a policy in simulation and then deploying it on a real system.
Work on transfer of sensorimotor control policies has mainly dealt with manual grasping and manipulation~\cite{gupta2017learning,wulfmeier2017mutual,bousmalis2018using, james2017transferring, rusu2016sim, Sadeghi_2018}.
In driving scenarios, synthetic data was mainly used to train perception systems for high-level tasks, such as semantic segmentation and object detection~\cite{Richter2016,johnson2017driving}. 
One exception is the work of M{\"{u}}ller et al.~\cite{muller2018driving}, which uses modularization to deploy a control policy learned in simulation on a physical ground vehicle.
Domain transfer has also been used for drone control: Sadeghi and Levine~\cite{cad2rl} learned a collision avoidance policy by using 3D simulation with extensive domain randomization.

Akin to many of the aforementioned methods, we use domain randomization~\cite{tobin2017domain} and modularization~\cite{muller2018driving} to increase generalization and achieve sim-to-real transfer.
Our work applies these techniques to drone racing. Specifically, we identify the most important factors for generalization and transfer with extensive analyses and ablation studies.

\section{Method}
\label{sec:methodology}

We address the problem of robust, agile flight of a quadrotor in a dynamic environment.
Our approach makes use of two subsystems: perception and control.
The perception system uses a Convolutional Neural Network (CNN) to predict a goal direction in local image coordinates, together with a desired navigation speed, from a single image collected by a forward-facing camera.
The control system uses the navigation goal produced by the perception system to generate a minimum-jerk trajectory~\cite{mueller2013computationally} that is tracked by a low-level controller~\cite{faessler2018differential}.
In the following, we describe the subsystems in more detail.

\textbf{Perception system.}\label{sec:perception_system}
The goal of the perception system is to analyze the image and provide a desired flight direction and navigation speed for the robot.
We implement the perception system by a convolutional network.
The network takes as input a $300\times200$ pixel RGB image, captured from the onboard camera, and outputs a tuple~$\lbrace \vec{x}, v \rbrace$, where $\vec{x} \in [-1,1]^2$ is a two-dimensional vector that encodes the direction to the new goal in normalized image coordinates, and $v \in [0,1]$ is a normalized desired speed to approach it.
To allow for onboard computing, we employ a modification of the DroNet architecture of Loquercio et al.~\cite{dronet}.
In section~\ref{sec:arch_search}, we will present the details of our architecture, which was designed to optimize the trade-off between accuracy and inference time.
With our hardware setup, the network achieves an inference rate of $15$ frames per second while running concurrently with the full control stack.
The system is trained by imitating an automatically computed expert policy, as explained in Section~\ref{sec:training_procedure}.

\textbf{Control system.}\label{sec:control_system}
Given the tuple $\lbrace \vec{x}, v \rbrace$, the control system generates low-level commands.
To convert the goal position $\vec{x}$ from two-dimensional normalized image coordinates to three-dimensional local frame coordinates,
we back-project the image coordinates $\vec{x}$ along the camera projection ray and derive the
goal point at a depth equal to the prediction horizon $d$ (see Figure~\ref{fig:traj_sketch}).
We found setting $d$ proportional to the normalized platform speed $v$ predicted by the network to work well.
The desired quadrotor speed $v_{des}$ is computed by rescaling the predicted normalized speed $v$ by a user-specified maximum speed $v_{max}$: $v_{des} = v_{max} \cdot v$.
This way, with a single trained network, the user can control the aggressiveness of flight by varying the maximum speed.
Once $p_g$ in the quadrotor's body frame and $v_{des}$ are available,
a state interception trajectory $t_s$ is computed to reach the goal position (see Figure~\ref{fig:traj_sketch}).
Since we run all computations onboard, we use computationally efficient minimum-jerk trajectories~\cite{mueller2013computationally} to generate $t_s$.
To track $t_s$, i.e.\ to compute the low-level control commands, we employ the control scheme proposed by Faessler et al.~\cite{faessler2018differential}.

\subsection{Training Procedure}\label{sec:training_procedure}

\begin{figure}
			\centering
			\includegraphics[width=0.9\linewidth]{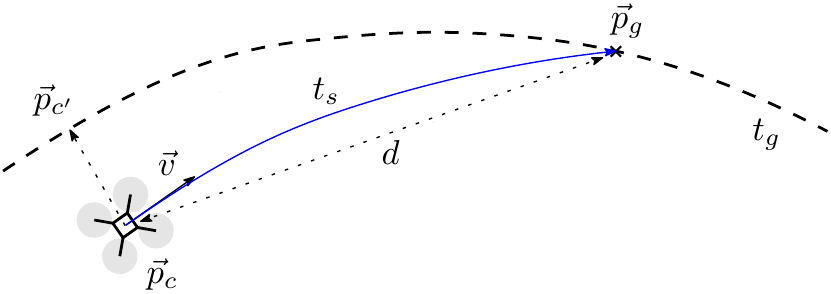}
			\caption{The pose $\vec{p}_c$ of the quadrotor is projected on the global trajectory $t_{g}$ to find the point $\vec{p}_{c'}$.
				The point at distance $d$ from the current quadrotor position $\vec{p}_c$, which belongs to $t_{g}$ in the forward direction with respect to $\vec{p}_{c'}$, defines the desired goal position $\vec{p}_{g}$.
			To push the quadrotor towards the reference trajectory $t_{g}$, a short trajectory segment $t_s$ is planned and tracked in a receding horizon fashion.}
\label{fig:traj_sketch}
\end{figure}

We train the perception system with imitation learning, using automatically generated globally optimal trajectories as a source of supervision.
To generate these trajectories, we make the assumption that at training time the location of each gate of the race track, expressed in a common reference frame, is known.
Additionally, we assume that at training time the quadrotor has access to accurate state estimates with respect to the latter reference frame.
Note however that at test time no privileged information is needed and the quadrotor relies on image data only.
The overall training setup is illustrated in Figure~\ref{fig:traj_sketch}.

\textbf{Expert policy.}
We first compute a global trajectory~$t_{g}$ that passes through all gates of the track, using the minimum-snap trajectory implementation from Mellinger and Kumar~\cite{mellinger2011minimum}.
To generate training data for the perception network, we implement an expert policy that follows the reference trajectory.
Given a quadrotor position $\vec{p}_c \in \mathbb{R}^3$, we compute the closest point~$\vec{p}_{c'} \in \mathbb{R}^3$ on the global reference trajectory.
The desired position $\vec{p}_{g} \in \mathbb{R}^3$ is defined as the point on the global reference trajectory the distance of which from $\vec{p}_c$ is equal to the prediction horizon $d \in \mathbb{R}$.
We project the desired position $\vec{p}_{g}$ onto the image plane of the forward facing camera to generate the ground truth normalized image coordinates $\vec{x}_{g}$ corresponding to the goal direction.
The desired speed $v_{g}$ is defined as the speed of the reference trajectory at~$\vec{p}_{c'}$ normalized by the maximum speed achieved along $t_g$.

\textbf{Data collection.}
 To train the network, we collect a dataset of state estimates and corresponding camera images.
 Using the global reference trajectory, we evaluate the expert policy on each of these samples and use the result as the ground truth for training.
 An important property of this training procedure is that it is agnostic to how exactly the training dataset is collected.
 We use this flexibility to select the most suitable data collection method when training in simulation and in the real world.
 The key consideration here is how to deal with the domain shift between training and test time.
 In our scenario, this domain shift mainly manifests itself when the quadrotor flies far from the reference trajectory $t_g$.
 In simulation, we employed a variant of DAgger~\cite{dagger}, which uses the expert policy to recover whenever the learned policy deviates far from the reference trajectory.
 Repeating the same procedure in the real world would be infeasible: allowing a partially trained network to control a UAV would pose a high risk of crashing and breaking the platform.
 Instead, we manually carried the quadrotor through the track and ensured a sufficient coverage of off-trajectory positions.

\textbf{Generating data in simulation.}
In our simulation experiment, we perform a modified version of DAgger~\cite{dagger} to train our flying policy.
On the data collected through the expert policy (Section~\ref{sec:training_procedure}) (in our case we let the expert policy fly for $\SI{40}{\second}$),
the network is trained for 10 epochs on the accumulated data. 
In the following run, the trained network is predicting actions, which are only executed if they keep the quadrotor within a margin $\epsilon$ from the global trajectory. 
In case the network's action violates this constraint, the expert policy is executed, generating a new training sample. 
This procedure is an automated form of DAgger~\cite{dagger} and allows the network to recover when deviating from the global trajectory. 
After another $\SI{40}{\second}$ of data generation, the network is retrained on all the accumulated data for 10 epochs. 
As soon as the network performs well on a given margin $\epsilon$, the margin is increased.
This process repeats until the network can eventually complete the whole track without help of the expert policy. 
In our simulation experiments, the margin $\epsilon$ was set to $\SI{0.5}{\meter}$ after the first training iteration.
The margin was incremented by $\SI{0.5}{\meter}$ as soon as the network could complete the track with limited help from the expert policy (less than 50 expert actions needed). 
For experiments on the static track, 20k images were collected, while for dynamic experiments 100k images of random gate positions were generated.

\textbf{Generating data in the real world.}
For safety reasons, it is not possible to apply DAgger for data collection in the real world.
Therefore, we ensure sufficient coverage of the possible actions by manually carrying the quadrotor through the track.
During this procedure, which we call \textit{handheld} mode, the expert policy is constantly generating training samples. 
Due to the drift of onboard state estimation, data is generated for a small part of the track before the quadrotor is reinitialized at a known position.
For the experiment on the static track, 25k images were collected, while for the dynamic experiment an additional 15k images were collected for different gate positions. 
For the narrow gap and occlusion experiments, 23k images were collected.

\textbf{Loss function.} We train the network with a weighted MSE loss on point and velocity predictions:
\begin{equation}
	L = \Vert \vec{x}  - \vec{x}_{g}\Vert^2 + \gamma (v - v_{g})^2,
	\label{eq:loss}
\end{equation}
where $\vec{x}_{g}$ denotes the groundtruth normalized image coordinates and $v_{g}$ denotes the groundtruth normalized speed.
By cross-validation, we found the optimal weight to be $\gamma=0.1$,
even though the performance was mostly insensitive to this parameter (see Appendix for details).

\textbf{Dynamic environments.}
The described training data generation procedure is limited to static environments, since the trajectory generation method is unable to take the changing geometry into account.
How can we use it to train a perception system that would be able to cope with dynamic environments?
Our key observation is that training on multiple static environments (for instance with varying gate positions) is sufficient to operate in dynamic environments at test time.
We collect data from multiple layouts generated by moving the gates from their initial position.
We compute a global reference trajectory for each layout and train a network jointly on all of these.
This simple approach supports generalization to dynamic tracks, with the additional benefit of improving the robustness of the system.

\textbf{Sim-to-real transfer.} One of the big advantages of perception-control modularization is that it allows training the perception block exclusively in simulation and then directly applying on the real system by leaving the control part unchanged.
As we will show in the experimental section, thanks to the abundance of simulated data, it is possible to train policies that are extremely robust to changes in environmental conditions, such as illumination, viewpoint, gate appearance, and background.
In order to collect diverse simulated data, we perform visual scene randomization in the simulated environment, while keeping the approximate track layout fixed. 
Apart from randomizing visual scene properties, the data collection procedure remains unchanged.

\begin{figure*}
\centering
\begin{subfigure}{.45\textwidth}
  \centering
  \includegraphics[width=0.9\linewidth]{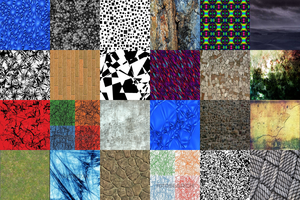}
  \caption{}
  \label{fig:train_bkgs}
\end{subfigure}%
\begin{subfigure}{.45\textwidth}
  \centering
  \includegraphics[width=0.9\linewidth]{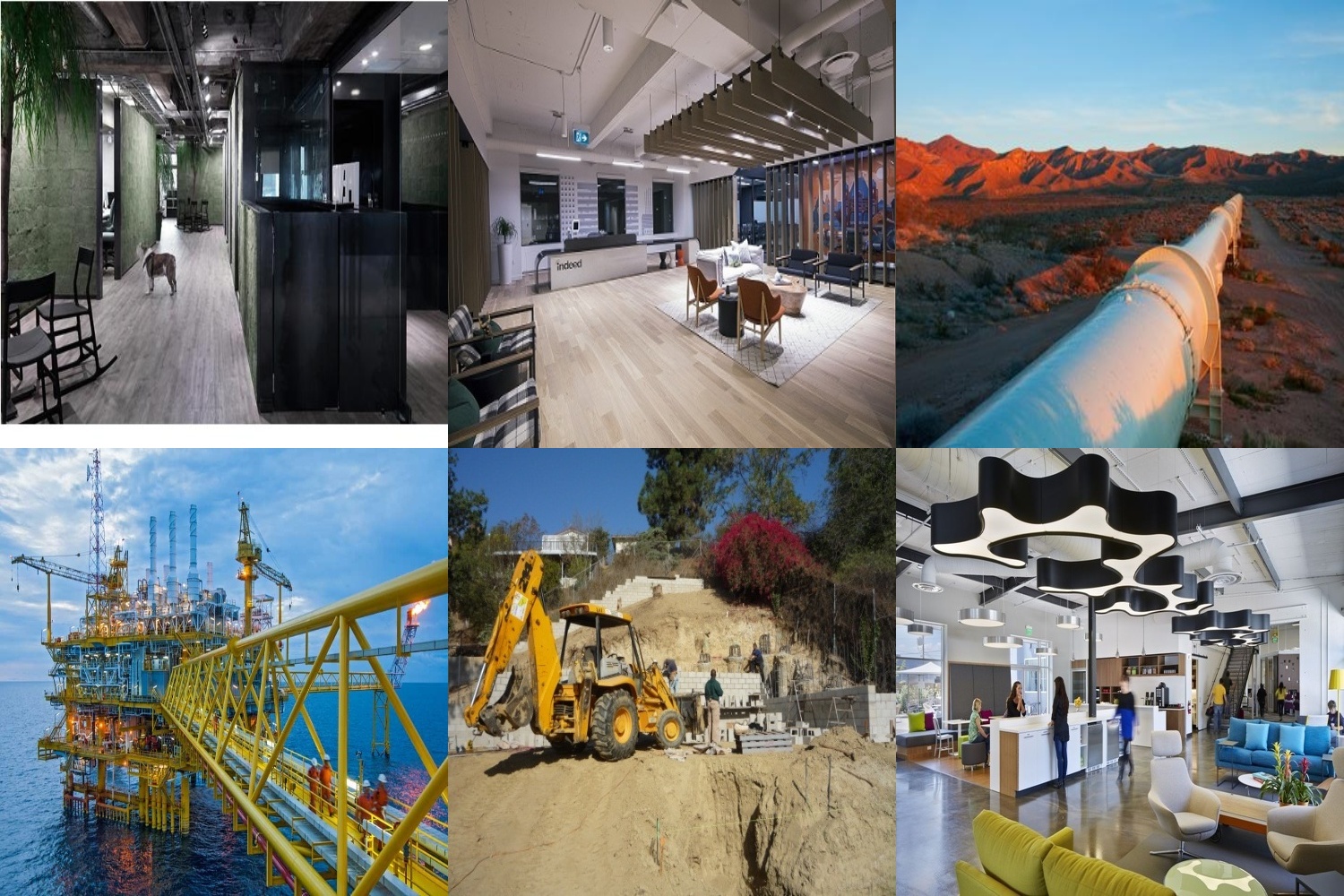}
  \caption{}
  \label{fig:test_bkgs}
\end{subfigure}
\par\bigskip 
\begin{subfigure}{.45\textwidth}
  \centering
  \includegraphics[width=0.9\linewidth]{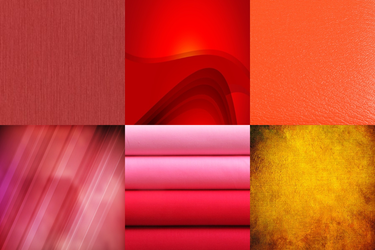}
  \caption{}
  \label{fig:gate_textures}
\end{subfigure}%
\begin{subfigure}{.45\textwidth}
  \centering
  \includegraphics[width=0.9\linewidth]{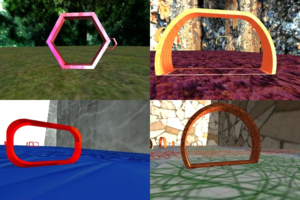}
  \caption{}
  \label{fig:gate_sim_examples}
\end{subfigure}
\caption{To test the generalization abilities of our approach, we randomize the visual properties of the environment (background, illumination, gate shape, and gate texture). This figure illustrates the random textures and shapes applied both at training (\textbf{a}) and test time(\textbf{b}). For space reasons, not all examples are shown. In total, we used 30 random backgrounds during training and 10 backgrounds during testing. We generated 6 different shapes of gates and used 5 of them for data generation and one for evaluation. Similarly, we used 10 random gate textures during training and a different one during evaluation. \textbf{a)} Random backgrounds used during training data generation. \textbf{b)} Random backgrounds used at test time. \textbf{c)} Gate textures. \textbf{d)} Selection of training examples illustrating the gate shapes and variation in illumination properties.}
\label{fig:sim_randomization}
\end{figure*}

We randomize the following visual scene properties: (i) the textures of the background, floor, and gates, (ii) the shape of the gates, and (iii) the lighting in the scene.
For (i), we apply distinct random textures to background and floor from a pool of 30 diverse synthetic textures~(Figure~\ref{fig:train_bkgs}).
The gate textures are drawn from a pool of 10 mainly red/orange textures (Figure~\ref{fig:gate_textures}).
For gate shape randomization (ii), we create 6 gate shapes of roughly the same size as the original gate.  
Figure~\ref{fig:gate_sim_examples} illustrates four of the different gate shapes used for data collection.
To randomize illumination conditions (iii), we perturb the ambient and emissive light properties of all textures (background, floor, gates). 
Both properties are drawn separately for background, floor, and gates from uniform distributions with support $[0, 1]$ for the ambient property and $[0, 0.3]$ for the emissive property.

While the textures applied during data collection are synthetic, the textures applied to background and floor at test time represent common indoor and outdoor environments (Figure~\ref{fig:test_bkgs}).
For testing we use held-out configurations of gate shape and texture not seen during training.

\subsection{Trajectory Generation}

\textbf{Generation of global trajectory.}
Both in simulation and in real-world experiments, a global trajectory is used to generate ground truth labels. 
To generate the trajectory, we use the implementation of Mellinger and Kumar~\cite{mellinger2011minimum}.
The trajectory is generated by providing a set of waypoints to pass through, a maximum velocity to achieve, as well as constraints on maximum thrust and body rates.
Note that the speed on the global trajectory is not constant. 
As waypoints, the centers of the gates are used. 
Furthermore, the trajectory can be shaped by additional waypoints, for example if it would pass close to a wall otherwise. 
In both simulation and real-world experiments, the maximum normalized thrust along the trajectory was set to $\SI{18}{\meter\per\square\second}$ and the maximum roll and pitch rate to $\SI{1.5}{\radian\per\second}$.
The maximum speed was chosen based on the dimensions of the track.
For the large simulated track, a maximum speed of $\SI{10}{\meter\per\second}$ was chosen, while on the smaller real-world track $\SI{6}{\meter\per\second}$.

\textbf{Generation of trajectory segments.}
The proposed navigation approach relies on constant recomputation of trajectory segments $t_s$ based on the output of a CNN. 
Implemented as state-interception trajectories, $t_s$ can be computed by specifying a start state, goal state and a desired execution time.
\cready{The velocity predicted by the network is used to compute the desired execution time of the trajectory segment $t_s$.}
While the start state of the trajectory segment is fully defined by the quadrotor's current position, velocity, and acceleration, the end state is only constrained by the goal position $p_g$, leaving velocity and acceleration in that state unconstrained.
This is, however, not an issue, since only the first part of each trajectory segment is executed in a receding horizon fashion.
Indeed, any time a new network prediction is available, a new state interception trajectory $t_s$ is calculated.

The goal position $p_g$ is dependent on the prediction horizon $d$ (see Section~\ref{sec:training_procedure}), which directly influences the aggressiveness of a maneuver.
\cready{Since the shape of the trajectory is only constrained by the start state and end state, reducing the prediction horizon decreases the lateral deviation from the straight-line connection of start state and end state but also leads to more aggressive maneuvers. 
Therefore, a long prediction horizon is usually required on straight and fast parts of the track, while a short prediction horizon performs better in tight turns and in proximity of gates.}
A long prediction horizon leads to a smoother flight pattern, usually required on straight and fast parts of the track.
Conversely, a short horizon performs more agile maneuvers, usually required in tight turns and in the proximity of gates.

The generation of the goal position $p_g$ differs from training to test time. 
At training time, the quadrotor's current position is projected onto the global trajectory and propagated by a prediction horizon $d_{train}$. 
At test time, the output of the network is back-projected along the camera projection ray by a planning length $d_{test}$.

\cready{At training time, we define the prediction horizon $d_{train}$ as a function of distance from the last gate and the next gate to be traversed:}
\begin{equation}
d_{train} = \max \left(d_{min}, \min \left(\Vert s_{last} \Vert , \Vert s_{next} \Vert\right)\right) \, ,
\label{eq:d_train}
\end{equation}
where $s_{last} \in \mathbb{R}^3$ and $s_{next} \in \mathbb{R}^3$ are the distances to the corresponding gates and~$d_{min}$ represents the minimum prediction horizon.
\cready{The minimum distance between the last and the next gate is used instead of only the distance to the next gate to avoid jumps in the prediction horizon after a gate pass.}
In our simulated track experiment, a minimum prediction horizon of ${d_{min} = \SI{1.5}{\meter}}$ was used, while for the real track we used ${d_{min} = \SI{1.0}{\meter}}$.

At test time, since the output of the network is a direction and a velocity, the length of a trajectory segment needs to be computed. 
To distinguish the length of trajectory segments at test time from the same concept at training time, we call it \textit{planning length} at test time.
The planning length of trajectory segments is computed based on the velocity output of the network (computation based on the location of the quadrotor with respect to the gates is not possible at test time since we do not have knowledge about gate positions).
The objective is again to adapt the planning length such that both smooth flight at high speed and aggressive maneuvers in tight turns are possible. 
We achieve this versatility by computing the planning length according to this linear function:
\begin{equation}
d_{test} = \min \left[d_{max}, \max \left(d_{min}, m_d v_{out}\right)\right] \, ,
\label{eq:d_test}
\end{equation}
where $m_d = \SI{0.6}{\second}$, $d_{min}=\SI{1.0}{\meter}$ and $d_{max}=\SI{2.0}{\meter}$ in our real-world experiments, and $m_d = \SI{0.5}{\second}$, $d_{min}=\SI{2.0}{\meter}$ and $d_{max}=\SI{5.0}{\meter}$ in the simulated track.

\section{Experiments}

We extensively evaluate the presented approach in a wide range of simulated and real scenarios.
We first use a controlled, simulated environment to test the main building blocks of our system, i.e.\ the convolutional architecture and the perception-control modularization.
Then, to show the ability of our approach to control real quadrotors, we perform a second set of experiments on a physical platform.
We compare our approach to state-of-the-art methods, as well as to human drone pilots of different skill levels.
We also demonstrate that our system achieves zero-shot simulation-to-reality transfer.
A policy trained on large amounts of cheap simulated data shows increased robustness against external factors, e.g.\ illumination and visual distractors, compared to a policy trained only with data collected in the real world.
Finally, we perform an ablation study to identify the most important factors that enable successful policy transfer from simulation to the real world.

\subsection{Experimental Setup}\label{sec:exp_setup}

For all our simulation experiments we use Gazebo as the simulation engine.
\cready{Although non-photorealistic, we have selected this engine since it models with high fidelity the physics of a quadrotor via the RotorS extension~\cite{Furrer2016}.}

Specifically, we simulate the AscTec Hummingbird multirotor, which is equipped with a forward-looking $300\times200$ pixels RGB camera.

The platform is spawned in a flying space of cubical shape with side length of 70 meters, which contains the experiment-specific race track.
The flying space is bounded by background and floor planes whose textures are randomized in the simulation experiments of Section~\ref{sec:sim2real}. 

The large simulated race track (Figure~\ref{img:sim_track_large}) is inspired by a real track used in international competitions.
We use this track layout for all of our experiments, except the comparison against end-to-end navigation policies.
The track is travelled in the same direction (clockwise or counterclockwise) at training and testing time.
We will release all code required to run our simulation experiments upon acceptance of this manuscript. 

For real-world experiments, except for the ones evaluating sim-to-real transfer, we collected data in the real world.
We used an in-house quadrotor equipped with an Intel UpBoard and a Qualcomm Snapdragon Flight Kit.
While the latter is used for visual-inertial odometry, the former represents the main computational unit of the platform.
The Intel UpBoard was used to run all the calculations required for flying, from neural network prediction to trajectory generation and tracking.

\subsection{Experiments in Simulation}

\label{sec:sim_experiments}
\begin{figure*}
\centering
\begin{subfigure}{.48\textwidth}
  \centering
  \includegraphics[width=0.9\linewidth]{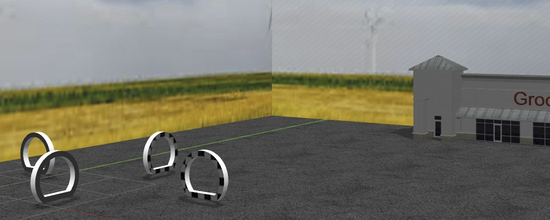}
  \caption{}
  \label{img:sim_track_small}
\end{subfigure}%
\begin{subfigure}{.48\textwidth}
  \centering
  \includegraphics[width=0.9\linewidth]{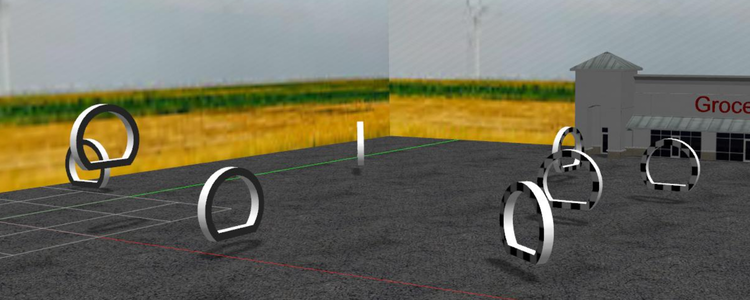}
  \caption{}
  \label{img:sim_track_large}
\end{subfigure}
\caption{Illustration of the simulated tracks. The small track (\textbf{a}) consists of 4 gates and spans a total length of 43 meters. The large track (\textbf{b}) consists of 8 gates placed at different heights and spans a total length of 116 meters.}
\end{figure*}

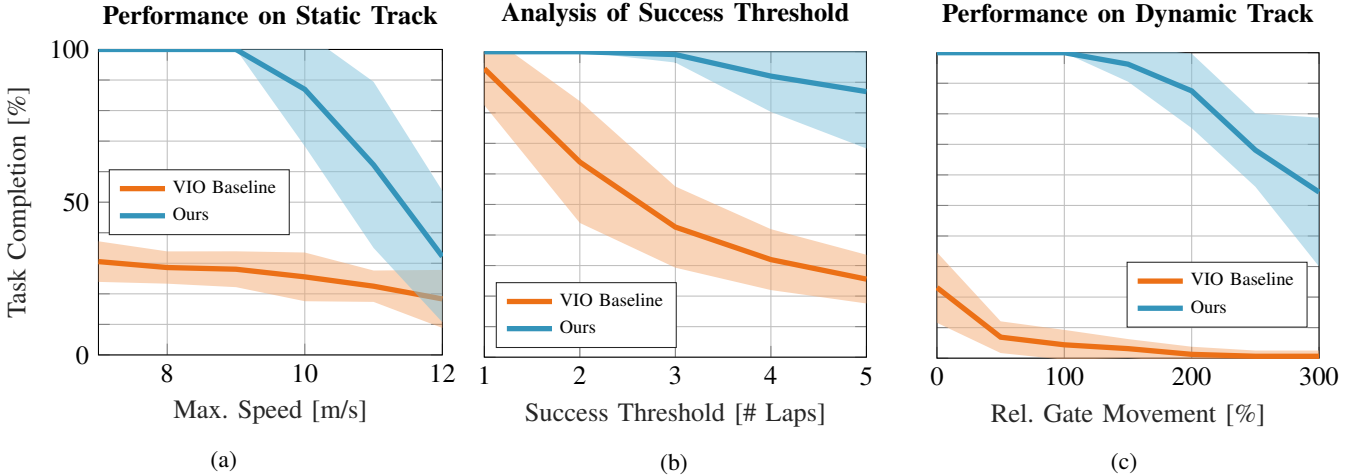
\begin{figure*}
\centering
\begin{subfigure}{.33\textwidth}
  \centering
%
%
\definecolor{mycolor1}{rgb}{0.95294,0.66275,0.44706}%
\definecolor{mycolor2}{rgb}{0.92549,0.43922,0.08627}%
\definecolor{mycolor3}{rgb}{0.50196,0.75686,0.85882}%
\definecolor{mycolor4}{rgb}{0.20392,0.58039,0.72941}%
\begin{tikzpicture}

\begin{axis}[%
width=1.8in,
height=1.6in,
at={(0.766in,0.486in)},
scale only axis,
xmin=7,
xmax=12,
xlabel style={font=\color{white!15!black}},
xlabel={Max. Speed [m/s]},
ymin=0,
ymax=100,
ytick={0,10,20,30,40,50,60,70,80,90,100,110},
yticklabels={0,,,,,50,,,,,100,},
ylabel style={font=\color{white!15!black}},
ylabel={Task Completion [\%]},
axis background/.style={fill=white},
title style={font=\bfseries},
title={Performance on Static Track},
xmajorgrids,
ymajorgrids,
legend style={at={(0.03,0.5)}, anchor=west, legend cell align=left, align=left, draw=white!15!black, font=\scriptsize}
]

\addplot[area legend, draw=none, fill=mycolor1, fill opacity=0.5, forget plot]
table[row sep=crcr] {%
x	y\\
7	37.1990337461674\\
8	33.8961937751223\\
9	33.9218644672064\\
10	33.5292778302263\\
11	27.6370116691408\\
12	27.8347952091595\\
12	8.83187145750718\\
11	17.3629883308592\\
10	17.5818332808848\\
9	22.1892466439048\\
8	23.3260284470999\\
7	23.9120773649437\\
}--cycle;
\addplot [color=mycolor2, line width=2.0pt]
  table[row sep=crcr]{%
7	30.5555555555556\\
8	28.6111111111111\\
9	28.0555555555556\\
10	25.5555555555556\\
11	22.5\\
12	18.3333333333333\\
};
\addlegendentry{VIO Baseline}

\addplot[area legend, draw=none, fill=mycolor3, fill opacity=0.5, forget plot]
table[row sep=crcr] {%
x	y\\
7	100\\
8	100\\
9	100\\
10	105.570060682489\\
11	89.4160850011565\\
12	53.7423821647184\\
12	10.702062279726\\
11	35.028359443288\\
10	68.3188282063998\\
9	100\\
8	100\\
7	100\\
}--cycle;
\addplot [color=mycolor4, line width=2.0pt]
  table[row sep=crcr]{%
7	100\\
8	100\\
9	100\\
10	86.9444444444444\\
11	62.2222222222222\\
12	32.2222222222222\\
};
\addlegendentry{Ours}

\end{axis}
\end{tikzpicture}%
  \caption{}
  \label{fig:comparison_max_speed}
\end{subfigure}%
\begin{subfigure}{.33\textwidth}
  \centering
%
%
\definecolor{mycolor1}{rgb}{0.95294,0.66275,0.44706}%
\definecolor{mycolor2}{rgb}{0.92549,0.43922,0.08627}%
\definecolor{mycolor3}{rgb}{0.50196,0.75686,0.85882}%
\definecolor{mycolor4}{rgb}{0.20392,0.58039,0.72941}%
\begin{tikzpicture}

\begin{axis}[%
width=2.0in,
height=1.6in,
at={(0.766in,0.486in)},
scale only axis,
xmin=1,
xmax=5,
xlabel style={font=\color{white!15!black}},
xlabel={Success Threshold [\# Laps]},
ymin=0,
ymax=100,
ytick={0,10,20,30,40,50,60,70,80,90,100,110},
yticklabels={\empty},
axis background/.style={fill=white},
title style={font=\bfseries},
title={Analysis of Success Threshold},
xmajorgrids,
ymajorgrids,
legend style={at={(0.03,0.03)}, anchor=south west, legend cell align=left, align=left, draw=white!15!black, font=\scriptsize}
]

\addplot[area legend, draw=none, fill=mycolor1, fill opacity=0.5, forget plot]
table[row sep=crcr] {%
x	y\\
1	106.392118426448\\
2	83.8231945755657\\
3	55.8821297170438\\
4	41.9115972877829\\
5	33.5292778302263\\
5	17.5818332808848\\
4	21.977291601106\\
3	29.3030554681414\\
2	43.954583202212\\
1	82.4967704624408\\
}--cycle;
\addplot [color=mycolor2, line width=2.0pt]
  table[row sep=crcr]{%
1	94.4444444444444\\
2	63.8888888888889\\
3	42.5925925925926\\
4	31.9444444444444\\
5	25.5555555555556\\
};
\addlegendentry{VIO Baseline}

\addplot[area legend, draw=none, fill=mycolor3, fill opacity=0.5, forget plot]
table[row sep=crcr] {%
x	y\\
1	100\\
2	100\\
3	101.692988078469\\
4	103.716874684967\\
5	105.570060682489\\
5	68.3188282063998\\
4	80.3109030928106\\
3	96.4551600696795\\
2	100\\
1	100\\
}--cycle;
\addplot [color=mycolor4, line width=2.0pt]
  table[row sep=crcr]{%
1	100\\
2	100\\
3	99.0740740740741\\
4	92.0138888888889\\
5	86.9444444444444\\
};
\addlegendentry{Ours}

\end{axis}
\end{tikzpicture}%
  \caption{}
  \label{fig:comparison_success_threshold}
\end{subfigure}%
\begin{subfigure}{.33\textwidth}
  \centering
%
%
\definecolor{mycolor1}{rgb}{0.95294,0.66275,0.44706}%
\definecolor{mycolor2}{rgb}{0.92549,0.43922,0.08627}%
\definecolor{mycolor3}{rgb}{0.50196,0.75686,0.85882}%
\definecolor{mycolor4}{rgb}{0.20392,0.58039,0.72941}%
\begin{tikzpicture}

\begin{axis}[%
width=2.0in,
height=1.6in,
at={(0.766in,0.486in)},
scale only axis,
xmin=0,
xmax=300,
xlabel style={font=\color{white!15!black}},
xlabel={Rel. Gate Movement [\%]},
ymin=0,
ymax=100,
ytick={0,10,20,30,40,50,60,70,80,90,100,110},
yticklabels={\empty},
ylabel style={font=\color{white!15!black}},
ylabel={\empty},
axis background/.style={fill=white},
title style={font=\bfseries},
title={Performance on Dynamic Track},
xmajorgrids,
ymajorgrids,
legend style={at={(0.97,0.1)}, anchor=south east, legend cell align=left, align=left, draw=white!15!black, font=\scriptsize}
]

\addplot[area legend, draw=none, fill=mycolor1, fill opacity=0.5, forget plot]
table[row sep=crcr] {%
x	y\\
0	34.6663658203871\\
50	12.0666399143238\\
100	9.25640604744166\\
150	6.25\\
200	3.75\\
250	2.5\\
300	2.5\\
300	-1.25\\
250	-1.25\\
200	-1.25\\
150	0\\
100	-0.506406047441659\\
50	1.6833600856762\\
0	11.5836341796129\\
}--cycle;
\addplot [color=mycolor2, line width=2.0pt]
  table[row sep=crcr]{%
0	23.125\\
50	6.875\\
100	4.375\\
150	3.125\\
200	1.25\\
250	0.625\\
300	0.625\\
};
\addlegendentry{VIO Baseline}

\addplot[area legend, draw=none, fill=mycolor3, fill opacity=0.5, forget plot]
table[row sep=crcr] {%
x	y\\
0	100\\
50	100\\
100	100\\
150	101.978219618695\\
200	99.6834929310112\\
250	80.1308579451866\\
300	78.75\\
300	30\\
250	56.1191420548134\\
200	75.3165070689888\\
150	90.5217803813052\\
100	100\\
50	100\\
0	100\\
}--cycle;
\addplot [color=mycolor4, line width=2.0pt]
  table[row sep=crcr]{%
0	100\\
50	100\\
100	100\\
150	96.25\\
200	87.5\\
250	68.125\\
300	54.375\\
};
\addlegendentry{Ours}

\end{axis}
\end{tikzpicture}%
  \caption{}
  \label{fig:comparison_moving_gates}
\end{subfigure}
\caption{\textbf{a)} Results of simulation experiments on the large track with static gates for different maximum speeds.
\textit{Task completion rate} measures the fraction of gates that were successfully completed without crashing.
A task completion rate of $100\%$ is achieved if the drone can complete five consecutive laps without crashing.
For each speed 10 runs were performed.
\textbf{b)} Analysis of the influence of the choice of success threshold. The experimental setting is the same as in Figure~\ref{fig:comparison_max_speed}, but the performance is reported for a fixed maximum speed of $\SI{10}{\meter\per\second}$ and different success thresholds.
The $y$-axis is shared with Figure~\ref{fig:comparison_max_speed}.
\textbf{c)} Result of our approach when flying through a simulated track with moving gates.
Every gate independently moves in a sinusoidal pattern with an amplitude proportional to its base size (\SI{1.3}{\meter}), with the indicated multiplier.
For each amplitude 10 runs were performed.
As for the static gate experiment, a task completion rate of $100\%$ is achieved if the drone can complete five consecutive laps without crashing.
Maximum speed is fixed to \SI{8}{\meter\per\second}.
The $y$-axis is shared with Figure~\ref{fig:comparison_max_speed}.
Lines denote mean performance, while the shaded areas indicate one standard deviation.
The reader is encouraged to watch the supplementary video to better understand the experimental setup and the task difficulty.}
\end{figure*}

Using a controlled simulated environment, we perform an extensive evaluation to (i) understand the advantages of our approach with respect to end-to-end or classical navigation policies, (ii) test the system's robustness to structural changes in the environment, and (iii) analyze the effect of the system's hyper-parameters on the final performance.

\textbf{Comparison to end-to-end learning approach.} In our first scenario, we use a small track that consists of four gates in a planar
configuration with a total length of 43 meters (Figure~\ref{img:sim_track_small}).

We use this track to compare the performance to a naive deep
learning baseline that directly regresses body rates from raw images.
Ground truth body rates for the baseline were provided
by generating a minimum snap reference trajectory through all gates and then tracking it with a low-level controller~\cite{faessler2018differential}.
For comparability, this baseline and our method share the same network architecture.
Our approach was always able to successfully complete the track. In contrast, the naive baseline
could never pass through more than one gate.
Training on more data (35K samples, as compared to 5K samples used by our method) did not noticeably improve the performance of the baseline.
We believe that end-to-end learning of low-level controls~\cite{mueller2017teaching} is suboptimal for the task of drone navigation when operating in the real world.
Since a quadrotor is an unstable platform~\cite{Narendra_1990}, learning the function that converts images to low-level commands has a very high sample complexity.
Additionally, the network is constrained by computation time. In order to guarantee stable control, the baseline network would have to produce control commands at a higher frequency (typically \SI{50}{\hertz}) than the camera images arrive (\SI{30}{\hertz}) and process them at a rate that is computationally infeasible with existing onboard hardware. In our experiments, since the low-level controller runs at \SI{50}{\hertz}, a network prediction is repeatedly applied until the next prediction arrives. 

In order to allow on-board sensing and computing, we propose a modularization scheme which organizes perception and control into two blocks.
With modularization, our approach can benefit from the most advanced learning based perceptual architectures and from years of study in the field of control theory~\cite{mahony2012multirotor}.
Because body rates are generated by a classic controller, the network
can focus on the navigation task, which leads to high sample efficiency.
Additionally, because the network does not need to ensure the stability of the platform,
it can process images at a lower rate than required for the low-level controller,
which unlocks onboard computation.
Given its inability to complete even this simple track, we do not conduct any further experiments with the direct end-to-end regression baseline.

\textbf{Performance on a complex track.}
In order to explore the capabilities of our approach of performing high-speed
racing, we conduct a second set of experiments on a larger and more complex track with 8 gates and a length of 116 meters (Figure~\ref{img:sim_track_large}).
The quantitative evaluation is conducted in terms of average task completion rate over five runs initialized with different random seeds.
For one run, the task completion rate linearly increases with each passed gate while 100\% task completion is achieved if the quadrotor is able to successfully complete five consecutive laps without crashing.
As a baseline, we use a pure feedforward setting by following the global trajectory $t_g$ using state estimates provided by visual inertial odometry~\cite{svo}.

The results of this experiment are shown in Figure~\ref{fig:comparison_max_speed}.
We can observe that the VIO baseline, due to accumulated drift, performs worse than our approach.
Figure~\ref{fig:comparison_success_threshold} illustrates the influence of drift on the baseline's performance.
While performance is comparable when one single lap is considered a success, it degrades rapidly if the threshold for success is raised to more laps.
\cready{On a static track (Figure~\ref{fig:comparison_max_speed}), a SLAM-based state estimator~\cite{orb_slam, qin2018vins} would have less drift than a VIO baseline, but we empirically found the latency of existing open-source SLAM pipelines to be too high for closed-loop control. A benchmark comparison of latencies of monocular visual-inertial SLAM algorithms for flying robots can be found in~\cite{delmericoICRA18}.}

Our approach works reliably up to a maximum speed of \SI{9}{\meter\per\second} and performance degrades gracefully at higher velocities.
The decrease in performance at higher speeds is mainly due to the higher body rates of the quadrotor that larger velocities inevitably entail.
Since the predictions of the network are in the body frame, the limited prediction frequency (\SI{30}Hz in the simulation experiments) is no longer sufficient to cope with the large roll and pitch rates of the platform at high velocities.

\textbf{Generalization to dynamic environments.} The learned policy has a characteristic that the expert policy lacks of: the ability to cope with dynamic environments.
\begin{figure}
    \centering
%
%
\definecolor{mycolor1}{rgb}{0.49400,0.18400,0.55600}%
\definecolor{mycolor2}{rgb}{0.92900,0.69400,0.12500}%
\definecolor{mycolor3}{rgb}{0.85000,0.32500,0.09800}%
\definecolor{mycolor4}{rgb}{0.00000,0.44700,0.74100}%
\begin{tikzpicture}

\begin{axis}[%
width=2.9in,
height=1.8in,
at={(0.758in,0.481in)},
scale only axis,
xmin=4,
xmax=12,
xlabel style={font=\color{white!15!black}\small},
xlabel={Speed [m/s]},
ymin=0,
ymax=100,
ylabel style={font=\color{white!15!black}\small},
ylabel={Task Completion [\%]},
axis background/.style={fill=white},
title style={font=\bfseries},
title={\empty},
xmajorgrids,
ymajorgrids,
legend style={at={(0.05,0.47)}, anchor=south west, legend cell align=left, align=left, draw=white!15!black, font=\scriptsize}
]

\addplot[area legend, draw=none, fill=mycolor1, fill opacity=0.2, forget plot]
table[row sep=crcr] {%
x	y\\
4	79\\
5	81\\
6	62\\
7	63\\
8	52\\
9	45\\
10	47\\
11	40\\
12	36\\
12	6\\
11	14\\
10	15\\
9	15\\
8	16\\
7	11\\
6	22\\
5	11\\
4	19\\
}--cycle;
\addplot [color=mycolor1, line width=1.5pt]
  table[row sep=crcr]{%
4	49\\
5	46\\
6	42\\
7	37\\
8	34\\
9	30\\
10	31\\
11	27\\
12	21\\
};
\addlegendentry{Background, Shape}

\addplot[area legend, draw=none, fill=mycolor2, fill opacity=0.2, forget plot]
table[row sep=crcr] {%
x	y\\
4	82\\
5	83\\
6	72\\
7	67\\
8	53\\
9	56\\
10	53\\
11	43\\
12	41\\
12	11\\
11	7\\
10	5\\
9	8\\
8	15\\
7	-1\\
6	16\\
5	13\\
4	20\\
}--cycle;
\addplot [color=mycolor2, line width=1.5pt]
  table[row sep=crcr]{%
4	51\\
5	48\\
6	44\\
7	33\\
8	34\\
9	32\\
10	29\\
11	25\\
12	26\\
};
\addlegendentry{Background, Illumination}

\addplot[area legend, draw=none, fill=mycolor3, fill opacity=0.2, forget plot]
table[row sep=crcr] {%
x	y\\
4	55\\
5	42\\
6	48\\
7	45\\
8	41\\
9	41\\
10	41\\
11	32\\
12	33\\
12	1\\
11	-2\\
10	3\\
9	-1\\
8	11\\
7	5\\
6	12\\
5	16\\
4	15\\
}--cycle;
\addplot [color=mycolor3, line width=1.5pt]
  table[row sep=crcr]{%
4	35\\
5	29\\
6	30\\
7	25\\
8	26\\
9	20\\
10	22\\
11	15\\
12	17\\
};
\addlegendentry{Background}

\addplot[area legend, draw=none, fill=mycolor4, fill opacity=0.2, forget plot]
table[row sep=crcr] {%
x	y\\
4	100\\
5	100\\
6	100\\
7	106.6\\
8	107.7\\
9	114\\
10	114\\
11	103\\
12	88\\
12	22\\
11	33\\
10	66\\
9	72\\
8	77.7\\
7	86.6\\
6	100\\
5	100\\
4	100\\
}--cycle;
\addplot [color=mycolor4, line width=1.5pt]
  table[row sep=crcr]{%
4	100\\
5	100\\
6	100\\
7	96.6\\
8	92.7\\
9	93\\
10	90\\
11	68\\
12	55\\
};
\addlegendentry{Background, Illumination, Shape}

\end{axis}
\end{tikzpicture}%
    \caption{Generalization tests on different backgrounds after domain randomization. 
    More comprehensive randomization increases the robustness of the learned policy to unseen scenarios at different speeds.
    Lines denote mean performance, while the shaded areas indicate one standard deviation.
    Background randomization has not been included in the analysis: without it the policy fails to complete even a single gate pass.
    }
    \label{fig:simulation_generalization}
\end{figure}
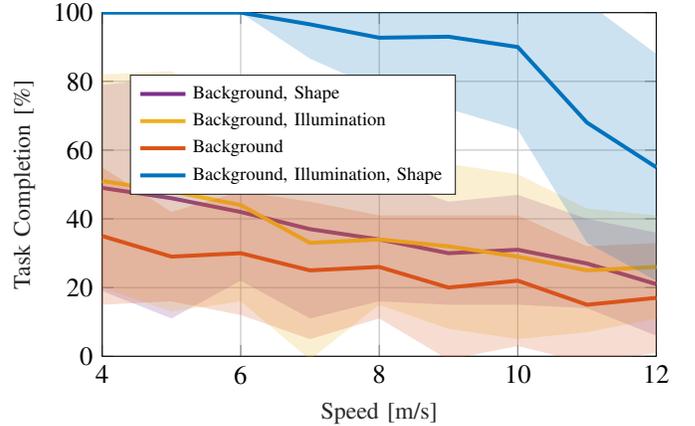
To quantitatively test this ability, we reuse the track layout from the previous experiment (Figure~\ref{img:sim_track_large}), but dynamically move each gate according to a sinusoidal pattern in each dimension independently.
Figure~\ref{fig:comparison_moving_gates} compares our system to the VIO baseline for varying amplitudes of gates' movement relative to their base size.
We evaluate the performance using the same metric as explained in Section~\ref{sec:sim_experiments}.
For this experiment, we kept the maximum platform velocity $v_{max}$ constant at \SI{8}{\meter\per\second}.
Despite the high speed, our approach can handle dynamic gate movements up to 1.5 times the gate diameter without crashing.
In contrast, the VIO baseline cannot adapt to changes in the environment, and fails even for small gate motions up to 50\% of the gate diameter.
The performance of our approach gracefully degrades for gate movements larger than 1.5 times the gate diameter, mainly due to the fact that consecutive gates get too close in flight direction while being shifted in other directions.
Such configurations require extremely sharp turns that go beyond the navigation capabilities of the system.
From this experiment, we can conclude that the proposed approach reactively adapts to dynamic changes in the environment and generalizes well to cases where the track layout remains roughly similar to the one used to collect training data.

\textbf{Generalization to changes in the simulation environment.} In the previous experiments, we have assumed a constant environment (background, illumination, gate shape) during data collection and testing.
In this section, we evaluate the generalization abilities of our approach to environment configurations not seen during training.
Specifically, we drastically change the environment background (Figure~\ref{fig:test_bkgs}) and use gate appearance and illumination conditions held out at training time.

Figure~\ref{fig:simulation_generalization} shows the result of this evaluation.
As expected, if data collection is performed in a single environment, the resulting policy has limited generalization (red line).
To make the policy environment-agnostic, we performed domain randomization while keeping the approximate track layout constant (details in Section~\ref{sec:training_procedure}).
Clearly, both randomization of gate shape and illumination lead to a policy that is more robust to new scenarios. 
Furthermore, while randomization of a single property leads to a modest improvement, performing all types of randomization simultaneously is crucial for good transfer.
Indeed, the simulated policy needs to be invariant to all of the randomized features in order to generalize well.

Surprisingly, as we show below, the learned policy can not only function reliably in simulation, but is also able to control a quadrotor in the real world.
In Section~\ref{sec:sim2real} we present an evaluation of the real world control abilities of this policy trained in simulation, as well as an ablation study to identify which of the randomization factors presented above are the most important for generalization and knowledge transfer.

\begin{figure}
\centering
\hspace*{-0.1in}
\includegraphics[scale=0.9]{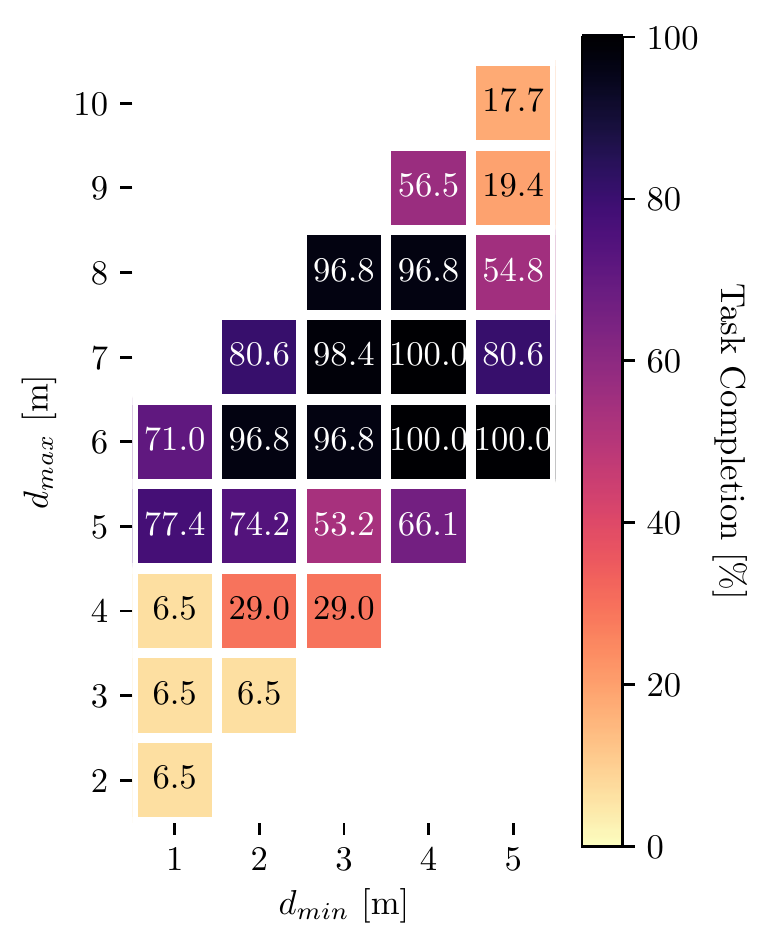}
\caption{\cready{Sensitivity analysis of planning length parameters $d_{min}$, $d_{max}$ on a simulated track. Maximum speed and (static) track layout are kept constant during the experiment.}}
\label{fig:sensitivity_planning_length}
\end{figure}

\cready{\textbf{Sensitivity to planning length.} 
We perform an ablation study of the \textit{planning length} parameters $d_{min}$, $d_{max}$ on a simulated track. Both the track layout and the maximum speed ($\SI{10.0}{\meter\per\second})$ are kept constant in this experiment. We varied $d_{\text{min}}$ between $\SI{1.0}{\meter}$ and $\SI{5.0}{\meter}$ and $d_{max}$ between $(d_{min} + 1.0)\SI{}{\meter}$ and $(d_{min} + 5.0)\SI{}{\meter}$. Figure~\ref{fig:sensitivity_planning_length} shows the results of this evaluation. For each configuration the average \textit{task completion rate} (Section~\ref{sec:sim_experiments}) over 5 runs is reported. Our systems performs well over a large range of $d_{min}$, $d_{max}$, with performance dropping sharply only for configurations with very short or very long planning lengths. This behaviour is expected, since excessively short planning lengths result in very aggressive maneuvers, while excessively long planning lengths restrict the agility of the platform.}

\subsection{Analysis of Accuracy and Efficiency}
\label{sec:arch_search}

The neural network at the core of our perception system constitutes the biggest computational bottleneck of our approach.
Given the constraints imposed by our processing unit, we can guarantee real-time performance only with relatively small CNNs.
Therefore, we investigated the relationship between the capacity (hence the representational power) of a neural network and its performance on the navigation task.
We measure performance in terms of both prediction accuracy on a validation set, and closed-loop control on a simulated platform, using, as above, completion rate as metric.
The capacity of the network is controlled through a multiplicative factor on the number of filters (in convolutional layers) and number of nodes (in fully connected layers).
The network with capacity $1.0$ corresponds to the DroNet architecture~\cite{dronet}.

\begin{figure}
\centering
%
%
\definecolor{mycolor1}{rgb}{0.00000,0.44700,0.74100}%
\definecolor{mycolor2}{rgb}{0.85000,0.32500,0.09800}%
\begin{tikzpicture}

\begin{axis}[%
width=2.5in,
height=1.8in,
at={(0.766in,0.486in)},
scale only axis,
xmin=0.25,
xmax=2,
xlabel style={font=\color{white!15!black}},
xlabel={Network Capacity},
separate axis lines,
every outer y axis line/.append style={mycolor1},
every y tick label/.append style={font=\color{mycolor1}},
every y tick/.append style={mycolor1},
ymin=0.14,
ymax=0.19,
ytick={0.14, 0.15,  0.16, 0.17, 0.18,  0.19},
ylabel style={font=\color{mycolor1}},
ylabel={Test Loss},
axis background/.style={fill=white},
xmajorgrids,
ymajorgrids,
legend style={legend cell align=left, align=left, draw=white!15!black}
]
\addlegendimage{/pgfplots/refstyle=plotyyref:leg0}
\addplot [color=mycolor1, thick]
  table[row sep=crcr]{%
0.25	0.18165902124585\\
0.5	0.173205080756888\\
1	0.167332005306815\\
1.5	0.161245154965971\\
2	0.158113883008419\\
};

\end{axis}

\begin{axis}[%
width=2.5in,
height=1.8in,
at={(0.766in,0.486in)},
scale only axis,
xmin=0.25,
xmax=2,
every outer y axis line/.append style={mycolor2},
every y tick label/.append style={font=\color{mycolor2}},
every y tick/.append style={mycolor2},
ymin=20,
ymax=120,
ytick={  0, 20, 40, 60, 80, 100, 120},
ylabel style={font=\color{mycolor2}},
ylabel={Inference Time [ms]},
axis y line*=right,
legend style={legend cell align=left, align=left, draw=white!15!black}
]
\addlegendimage{/pgfplots/refstyle=plotyyref:leg0}
\addplot [color=mycolor2, thick]
  table[row sep=crcr]{%
0.25	25.75\\
0.5	30.847\\
1	65.126\\
1.5	77.215\\
2	102.655\\
};

\end{axis}
\end{tikzpicture}%
\caption{Test loss and inference time for different network capacity factors. Inference time is measured on the actual platform. }
\label{fig:testloss_vs_time}
\end{figure}
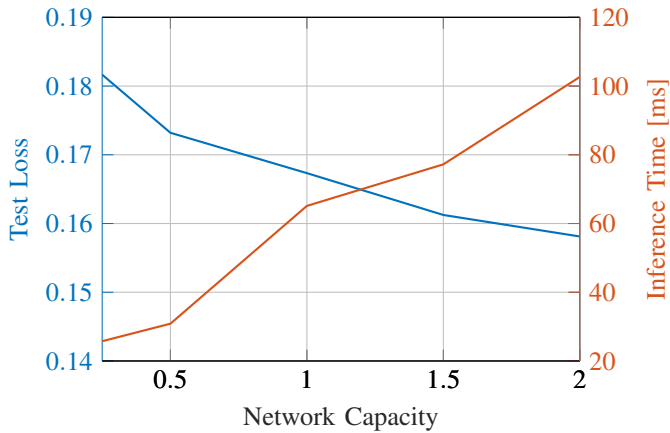

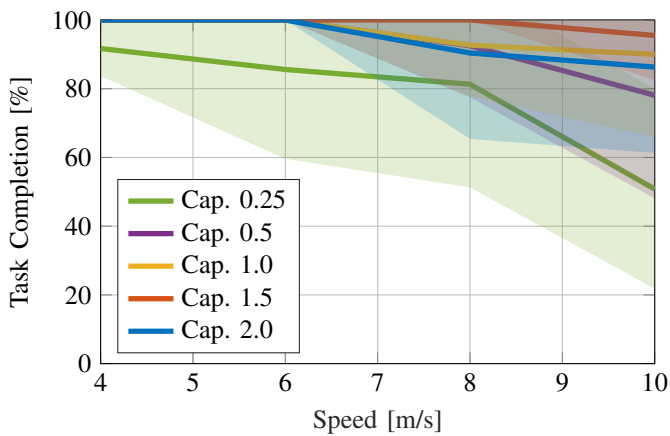
\begin{figure}
\centering
%
%
\definecolor{mycolor1}{rgb}{0.46600,0.67400,0.18800}%
\definecolor{mycolor2}{rgb}{0.49400,0.18400,0.55600}%
\definecolor{mycolor3}{rgb}{0.92900,0.69400,0.12500}%
\definecolor{mycolor4}{rgb}{0.85000,0.32500,0.09800}%
\definecolor{mycolor5}{rgb}{0.00000,0.44700,0.74100}%
\begin{tikzpicture}
\begin{axis}[%
width=2.9in,
height=1.8in,
at={(0.166in,0.486in)},
scale only axis,
xmin=4,
xmax=10,
xlabel style={font=\color{white!15!black}},
xlabel={Speed [m/s]},
ylabel={Task Completion [\%]},
ymin=0,
ymax=100,
axis background/.style={fill=white},
xmajorgrids,
ymajorgrids,
legend style={at={(0.03,0.03)}, anchor=south west, legend cell align=left, align=left, draw=white!15!black}
]

\addplot[area legend, draw=none, fill=mycolor1, fill opacity=0.2, forget plot]
table[row sep=crcr] {%
x	y\\
4	99.7\\
6	111.6\\
8	111.3\\
10	79.8\\
10	21.8\\
8	51.3\\
6	59.6\\
4	83.7\\
}--cycle;
\addplot [color=mycolor1, line width=2.0pt]
  table[row sep=crcr]{%
4	91.7\\
6	85.6\\
8	81.3\\
10	50.8\\
};
\addlegendentry{Cap. 0.25}

\addplot[area legend, draw=none, fill=mycolor2, fill opacity=0.2, forget plot]
table[row sep=crcr] {%
x	y\\
4	100\\
6	100\\
8	107.5\\
10	108.1\\
10	48.1\\
8	77.5\\
6	100\\
4	100\\
}--cycle;
\addplot [color=mycolor2, line width=2.0pt]
  table[row sep=crcr]{%
4	100\\
6	100\\
8	92.5\\
10	78.1\\
};
\addlegendentry{Cap. 0.5}

\addplot[area legend, draw=none, fill=mycolor3, fill opacity=0.2, forget plot]
table[row sep=crcr] {%
x	y\\
4	100\\
6	100\\
8	107.7\\
10	114\\
10	66\\
8	77.7\\
6	100\\
4	100\\
}--cycle;
\addplot [color=mycolor3, line width=2.0pt]
  table[row sep=crcr]{%
4	100\\
6	100\\
8	92.7\\
10	90\\
};
\addlegendentry{Cap. 1.0}

\addplot[area legend, draw=none, fill=mycolor4, fill opacity=0.2, forget plot]
table[row sep=crcr] {%
x	y\\
4	100\\
6	100\\
8	100\\
10	108.5\\
10	82.5\\
8	100\\
6	100\\
4	100\\
}--cycle;
\addplot [color=mycolor4, line width=2.0pt]
  table[row sep=crcr]{%
4	100\\
6	100\\
8	100\\
10	95.5\\
};
\addlegendentry{Cap. 1.5}

\addplot[area legend, draw=none, fill=mycolor5, fill opacity=0.2, forget plot]
table[row sep=crcr] {%
x	y\\
4	100\\
6	100\\
8	115.4\\
10	111.3\\
10	61.3\\
8	65.4\\
6	100\\
4	100\\
}--cycle;
\addplot [color=mycolor5, line width=2.0pt]
  table[row sep=crcr]{%
4	100\\
6	100\\
8	90.4\\
10	86.3\\
};
\addlegendentry{Cap. 2.0}

\end{axis}

\end{tikzpicture}%
\caption{Comparison of different network capacities on different backgrounds after domain randomization.}
\label{fig:closed_loop}
\end{figure}

Figure~\ref{fig:testloss_vs_time} shows the relationship between the network capacity, its test loss (RMSE) on a validation set, and its inference time on an Intel UpBoard (our onboard processing unit).
Given their larger parametrization, \cready{wider} architectures have a lower generalization error but largely increase the computational and memory budget required for their execution.
Interestingly, a lower generalization loss does not always correspond to a better closed-loop performance.
This can be observed in Figure~\ref{fig:closed_loop}, where the network with capacity $1.5$ outperforms the one with capacity $2.0$ at high speeds.
Indeed, as shown in Figure~\ref{fig:testloss_vs_time}, larger networks entail smaller inference rates, which result in a decrease in agility.

In our previous conference paper~\cite{kaufmann2018deep}, we used a capacity factor of $1.0$, which appears to have a good time-accuracy trade-off.
However, in the light of this study, we select a capacity factor of $0.5$ for all our new sim-to-real experiments to ease the computational burden.
Indeed, the latter experiments are performed at a speed of \SI{2}{\meter\per\second}, where both $0.5$ and $1.0$ have equivalent closed-loop control performance (Figure~\ref{fig:closed_loop}).

\subsection{Experiments in the Real World}

To show the ability of our approach to function in the real world, we performed experiments on a physical quadrotor.
We compared our model to state-of-the-art classic approaches to robot navigation, as well as to human drone pilots of different skill levels.

\begin{figure}
\centering
\includegraphics[width=0.9\linewidth]{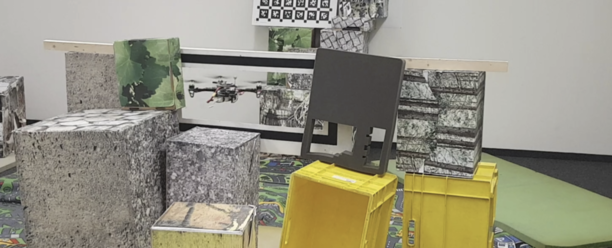}
\caption{Setup of the narrow gap and occlusion experiments.}
\label{img:narrow_gap}
\end{figure}

\textbf{Narrow gate passing.} In the initial set of experiments the quadrotor was required to pass through a  narrow gate, only slightly larger than the platform itself. These experiments are designed to test the robustness and precision of the proposed approach.
An illustration of the setup is shown in Figure~\ref{img:narrow_gap}.
We compare our approach to the handcrafted window detector of Falanga et al.~\cite{aggressive_falanga} by replacing our perception system with the handcrafted detector and leaving the control system unchanged.

\begin{table}
	\centering
	\begin{tabular}{c | c c}
		\toprule
		Relative Angle Range [$\degree$] & Handcrafted Detector & Network \\
		\midrule
		$[0, 30]$ & $70\%$ & $100\%$ \\
		$[30, 70]$ & $0\%$ & $80\%$ \\
		$[70, 90]$* & $0\%$ & $20\%$ \\
		\bottomrule
	\end{tabular}
	\caption{Success rate for flying through a narrow gap from different initial angles. Each row reports the average of ten runs uniformly spanning the range. The gate was completely invisible at initialization in the experiments marked with *.}
	\label{tab:static_window}
\end{table}

Table~\ref{tab:static_window} shows a comparison between our approach and the baseline. We tested the robustness of both approaches to the initial position of the quadrotor by placing the platform at different starting angles with respect to the gate (measured as the angle between the line joining the center of gravity of the quadrotor and the gate, respectively, and the optical axis of the forward facing camera on the platform).
We then measured the average success rate at passing the gate without crashing.
The experiments indicate that our approach is not sensitive to the initial position of the quadrotor. 
The drone is able to pass the gate consistently, even if the gate is only partially visible.
In contrast, the baseline sometimes fails even if the gate is fully visible because the window detector loses tracking due to platform vibrations.
When the gate is not entirely in the field of view, the handcrafted detector fails in all cases.

In order to further highlight the robustness and generalization abilities of the approach, we perform experiments with an increasing amount of clutter that occludes the gate.
Note that the learning approach has not been trained on such occluded configurations.
Figure~\ref{fig:comparison_occlusion} shows that our approach is robust to occlusions
of up to 50\% of the total area of the gate (Figure~\ref{img:narrow_gap}), whereas the handcrafted baseline
breaks down even for moderate levels of occlusion.
For occlusions larger than 50\% we observe a rapid drop in performance. This can be explained by the fact that the remaining gap was barely larger than the drone itself, requiring very high precision to successfully pass it. Furthermore, visual ambiguities of the gate itself become problematic. If just one of the edges of the window is visible, it is impossible to differentiate between the top and bottom part. This results in over-correction when the drone is very close to the gate.

\begin{figure}
\centering
%
%
\definecolor{mycolor1}{rgb}{0.00000,0.44700,0.74100}%
\definecolor{mycolor2}{rgb}{0.85000,0.32500,0.09800}%
\begin{tikzpicture}

\begin{axis}[%
width=2.5in,
height=1.2in,
at={(0.758in,0.481in)},
scale only axis,
xmin=0,
xmax=75,
xlabel style={font=\color{white!15!black}},
xlabel={Occlusion of Gate [\%]},
ymin=0,
ymax=100,
ylabel style={font=\color{white!15!black}},
ylabel={Succ. Gate Passes [\%]},
axis background/.style={fill=white},
title style={font=\bfseries},
title={\empty},
xmajorgrids,
ymajorgrids,
legend style={at={(0.03,0.1)}, anchor=south west, legend cell align=left, align=left, draw=white!15!black, font=\scriptsize}
]
\addplot [color=mycolor1, mark=asterisk, mark options={solid, mycolor1}, thick]
  table[row sep=crcr]{%
0	100\\
15	100\\
30	100\\
45	100\\
60	75\\
75	10\\
};
\addlegendentry{Ours}

\addplot [color=mycolor2, mark=square, mark options={solid, mycolor2}, thick]
  table[row sep=crcr]{%
0	70\\
15	0\\
30	0\\
45	0\\
60	0\\
75	0\\
};
\addlegendentry{Baseline}

\end{axis}
\end{tikzpicture}%
\caption{Success rate for different amounts of occlusion of the gate.
Our method is much more robust than the baseline method that makes use of a hand-crafted window detector.
Note that at more than $60\%$ occlusion, the platform has barely any space to pass through the gap.}
\label{fig:comparison_occlusion}
\end{figure}
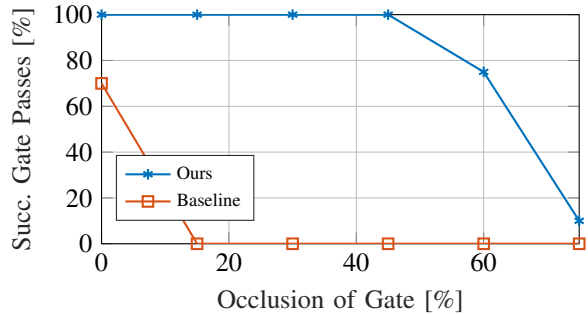

\begin{figure}
\centering
%
%
\definecolor{mycolor1}{rgb}{0.00000,0.44700,0.74100}%
\definecolor{mycolor2}{rgb}{0.85000,0.32500,0.09800}%
\definecolor{mycolor3}{rgb}{0.92900,0.69400,0.12500}%
\definecolor{mycolor4}{rgb}{0.49400,0.18400,0.55600}%
\definecolor{mycolor5}{rgb}{0.46600,0.67400,0.18800}%
\definecolor{mycolor6}{rgb}{0.30100,0.74500,0.93300}%
\definecolor{mycolor7}{rgb}{0.63500,0.07800,0.18400}%
\begin{tikzpicture}

\begin{axis}[%
width=1.6in,
height=1.8in,
at={(0.766in,0.486in)},
scale only axis,
xmin=20,
xmax=100,
xlabel style={font=\color{white!15!black}},
xlabel={Success Rate [\%]},
ymin=4,
ymax=18,
ylabel style={font=\color{white!15!black}},
ylabel={Best Lap Time [s]},
axis background/.style={fill=white},
axis x line*=bottom,
axis y line*=left,
xmajorgrids,
ymajorgrids,
legend style={at={(1.1,0.23)}, anchor=south west, legend cell align=left, align=left, draw=white!15!black, font=\scriptsize}
]
\addplot[only marks, mark=*, mark options={}, mark size=4pt, color=mycolor1, fill=mycolor1] table[row sep=crcr]{%
x	y\\
100	15.8\\
};
\addlegendentry{Ours [1m/s]}

\addplot[only marks, mark=*, mark options={}, mark size=4pt, color=mycolor2, fill=mycolor2] table[row sep=crcr]{%
x	y\\
100	13.4\\
};
\addlegendentry{Ours [2m/s]}

\addplot[only marks, mark=*, mark options={}, mark size=4pt, color=mycolor3, fill=mycolor3] table[row sep=crcr]{%
x	y\\
95	11.8\\
};
\addlegendentry{Ours [3m/s]}

\addplot[only marks, mark=diamond*, mark options={}, mark size=4pt, color=mycolor4, fill=mycolor4] table[row sep=crcr]{%
x	y\\
43.8	16.1\\
};
\addlegendentry{VIO [1m/s]}

\addplot[only marks, mark=diamond*, mark options={}, mark size=4pt, color=mycolor5, fill=mycolor5] table[row sep=crcr]{%
x	y\\
25	13.9\\
};
\addlegendentry{VIO [2m/s]}

\addplot[only marks, mark=square*, mark options={}, mark size=3pt, color=mycolor6, fill=mycolor6] table[row sep=crcr]{%
x	y\\
90	5\\
};
\addlegendentry{Professional Pilot}

\addplot[only marks, mark=square*, mark options={}, mark size=3pt, color=mycolor7, fill=mycolor7] table[row sep=crcr]{%
x	y\\
50	8.4\\
};
\addlegendentry{Intermediate Pilot}

\end{axis}
\end{tikzpicture}%
\caption{Results on a real race track composed of 4 gates.
Our learning-based approach compares favorably against a set of baselines based on visual-inertial state estimation.
Additionally, we compare against an intermediate and a professional human pilot.
We evaluate success rate using the same metric as explained in Section~\ref{sec:sim_experiments}.}
\label{fig:comparison_real_world}
\end{figure}
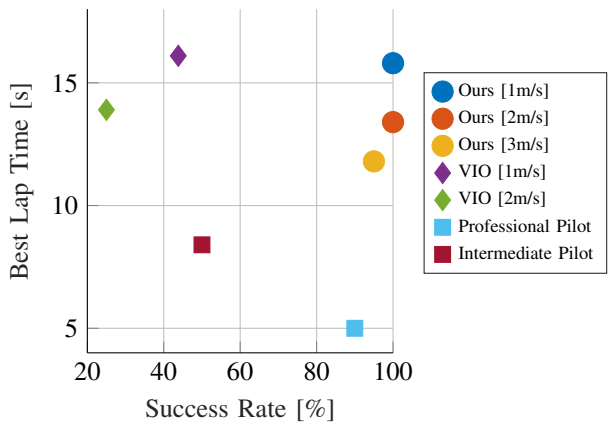

\begin{figure}
\centering
\includegraphics[width=0.9\linewidth]{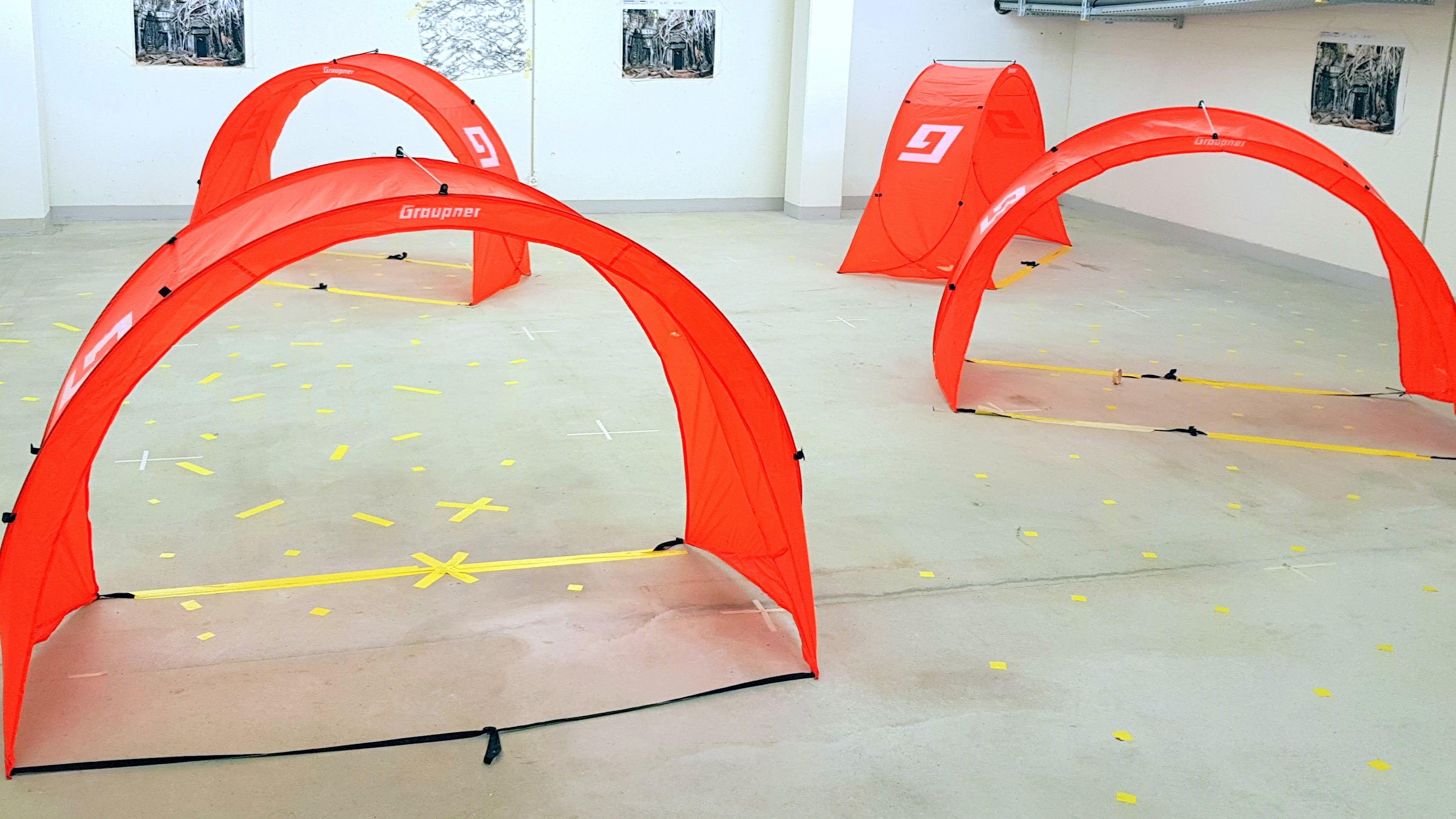}
\caption{Track configuration used for the real world experiments.}
\label{fig:real_track}
\end{figure}

\textbf{Experiments on a race track.}\label{sec:exp_real_track}
To evaluate the performance of our approach in a multi-gate scenario, we challenge the system to race through a track with either static or dynamic gates.
The track is shown in Figure~\ref{fig:real_track}. It is composed of four gates
and has a total length of 21~meters.

To fully understand the potential and limitations of our approach, we compared to a number of baselines, such as a classic approach based on planning and tracking~\cite{LoiBruMcGKum17} and human pilots of different skill levels.
Note that due to the smaller size of the real track compared to the simulated one, the maximum speed achieved in the real world experiments is lower than in simulation.
For our baseline, we use a state-of-the-art visual-inertial odometry (VIO) approach~\cite{LoiBruMcGKum17} for state estimation in order to track the global reference trajectory.

Figure~\ref{fig:comparison_real_world} summarizes the quantitative results of our evaluation, where we measure success rate (completing five consecutive laps without crashing corresponds to 100\%), as well as the best lap time.
Our learning-based approach outperforms the VIO baseline, whose drift at high speeds inevitably leads to poor performance.
In contrast, our approach is insensitive to state estimation drift, since it generates navigation commands in the body frame.
As a result, it completes the track with higher robustness and speed than the VIO baseline.

\begin{table}[b]
	\centering
	\begin{tabular}{c | c c | c c}
		\toprule
		& \multicolumn{2}{c|}{Task Completion (Average)} & \multicolumn{2}{c}{Best lap time [s]}\\
		Method & static & dynamic & static & dynamic\\
		\midrule
		Ours & 95\% & 95\% & 12.1 & 15.0\\
		Professional Pilot & 90\% & 80\% & 5.0 & 6.5 \\
		\bottomrule
	\end{tabular}
	\caption{Comparison of our approach with a professional human pilot on a static and a dynamic track. We evaluate the performance using the same metric as explained in Section~\ref{sec:sim_experiments}.}
	\label{tab:static_dynamic_comparison}
\end{table}

In order to see how state-of-the-art autonomous approaches compare to human pilots, we asked a professional and an intermediate pilot to race through the track in first-person view. 
We allowed the pilots to practice the track for 10 laps before lap times and failures were measured (Table~\ref{tab:static_dynamic_comparison}).
It is evident from Figure~\ref{fig:comparison_real_world} that both the professional and the intermediate pilots were able to complete the track faster than the autonomous systems. 
However, the high speed and aggressive flight by human pilots comes at the cost of increased failure rates. 
The intermediate pilot in particular had issues with the sharp turns present in the track, leading to frequent crashes.
Compared with the autonomous systems, human pilots perform more agile maneuvers, especially in sharp turns.
Such maneuvers require a level of reasoning about the environment that our autonomous system still lacks.

\textbf{Dynamically moving gates.} We performed an additional experiment to understand the abilities of our approach to adapt to dynamically changing environments.
In order to do so, we manually moved the gates of the race track (Figure~\ref{fig:real_track}) while the quadrotor was navigating through it.
Flying the track under these conditions requires the navigation system to reactively respond to dynamic changes.
Note that moving gates break the main assumption of traditional high-speed navigation approaches~\cite{mit_plan, teach_repeat}, specifically that the trajectory can be pre-planned in a static world. They could thus not be deployed in this scenario.
Due to the dynamic nature of this experiment, we encourage the reader to watch the supplementary video\footnote{Available from: \url{http://youtu.be/8RILnqPxo1s}}.
Table~\ref{tab:static_dynamic_comparison} provides a comparison in term of task completion and lap time with respect to a professional pilot.
Due to the gates' movement, lap times are larger than the ones recorded in static conditions.
However, while our approach achieves the same performance with respect to crashes, the human pilot performs slightly worse, given the difficulties entailed by the unpredictability of the track layout.
It is worth noting that training data for our policy was collected by changing the position of only a single gate, but the network was able to cope with movement of any gate at test time.

\subsection{Simulation to Real World Transfer}\label{sec:sim2real}

We now attempt direct simulation-to-real transfer of the navigation system.
To train the policy in simulation, we use the same process to collect simulated data as in Section~\ref{sec:sim_experiments}, i.e. randomization of illumination conditions, gate appearance, and background.
The resulting policy, evaluated in simulation in Figure~\ref{fig:simulation_generalization}, is then used without any finetuning to fly a real quadrotor.
Despite the large appearance differences between the simulated environment (Figure~\ref{fig:gate_sim_examples}) and the real one (Figure~\ref{fig:real_track}), the policy trained in simulation via domain randomization has the ability to control the quadrotor in the real world.
Thanks to the abundance of simulated data, this policy can not only be transferred from simulation to the real world, but is also more robust to changes in the environment than the policy trained with data collected on the real track.
As can be seen in the supplementaty video, the policy learned in simulation can not only reliably control the platform, but is also robust to drastic differences in illumination and distractors on the track.

\begin{figure}
\centering
%
%
\definecolor{mycolor1}{rgb}{0.00000,0.44700,0.74100}%
\definecolor{mycolor2}{rgb}{0.85000,0.32500,0.09800}%
\begin{tikzpicture}

\begin{axis}[%
width=2.9in,
height=1.8in,
at={(0.758in,0.481in)},
scale only axis,
bar shift auto,
xmin=0.514285714285714,
xmax=3.48571428571429,
xtick={1,2,3},
xticklabels={{Easy},{Medium},{Difficult}},
xlabel style={font=\color{white!15!black}},
xlabel={Illumination},
ymin=0,
ymax=100,
ylabel style={font=\color{white!15!black}},
ylabel={Task Completion [\%]},
axis background/.style={fill=white},
xmajorgrids,
ymajorgrids,
legend style={legend cell align=left, align=left, draw=white!15!black, font=\scriptsize}
]
\addplot[ybar, bar width=0.229, fill=mycolor1, draw=black, area legend] table[row sep=crcr] {%
1	100\\
2	95\\
3	75\\
};
\addplot[forget plot, color=white!15!black] table[row sep=crcr] {%
0.514285714285714	0\\
3.48571428571429	0\\
};
\addlegendentry{Sim2Real}

\addplot[ybar, bar width=0.229, fill=mycolor2, draw=black, area legend] table[row sep=crcr] {%
1	100\\
2	100\\
3	53\\
};
\addplot[forget plot, color=white!15!black] table[row sep=crcr] {%
0.514285714285714	0\\
3.48571428571429	0\\
};
\addlegendentry{Real}

\end{axis}
\end{tikzpicture}%
\caption{Performance comparison (measured with task completion rate) of the model trained in simulation and the one trained with real data. With \textit{easy} and \textit{medium} illumination (on which the real model was trained on), the approaches achieve comparable performance. However, with \textit{difficult} illumination the simulated model outperforms the real one, since the latter was never exposed to this degree of illumination changes at training time. The supplementary video illustrates the different illumination conditions. }
\label{fig:sim_vs_real}
\end{figure}
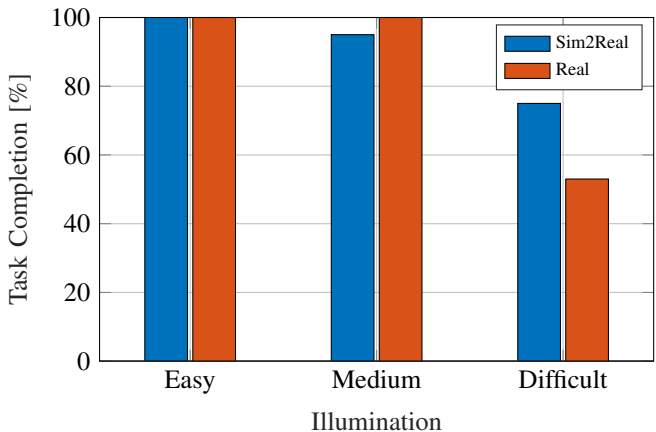

To quantitatively benchmark the policy learned in simulation, we compare it against a policy that was trained on real data. 
We use the same metric as explained in Section~\ref{sec:sim_experiments} for this evaluation.
All experiments are repeated $10$ times and the results averaged.
The results of this evaluation are shown in Figure~\ref{fig:sim_vs_real}.
The data that was used to train the ``real'' policy was recorded on the same track for two different illumination conditions, \textit{easy} and \textit{medium}.
Illumination conditions are varied by changing the number of enabled light sources: $4$ for the easy, $2$ for the medium, and $1$ for the difficult.
The supplementary video illustrates the different illumination conditions. 

The policy trained in simulation performs on par with the one trained with real data in experiments that have the same illumination conditions as the training data of the real policy. 
However, when the environment conditions are drastically different (i.e.\ with very challenging illumination) the policy trained with real data is outperformed by the one trained in simulation.
Indeed, as shown by previous work~\cite{james2017transferring}, the abundance of simulated training data makes the resulting learning policy robust to environmental changes.
We invite the reader to watch the supplementary video to understand the difficulty of this last set of experiments.

\textbf{What is important for transfer?} 
We conducted a set of ablation studies to understand what are the most important factors for transfer from simulation to the real world.
In order to do so, we collected a dataset of real world images from both indoor and outdoor environments in different illumination conditions, which we then annotated using the same procedure as explained in Section~\ref{sec:methodology}.
More specifically, the dataset is composed of approximately $10$K images and is collected from 3 indoor environments under different illumination conditions.
Sample images of this dataset are shown in the appendix.

\begin{figure}
\centering
\hspace*{-0.1in}
\includegraphics[scale=1.0]{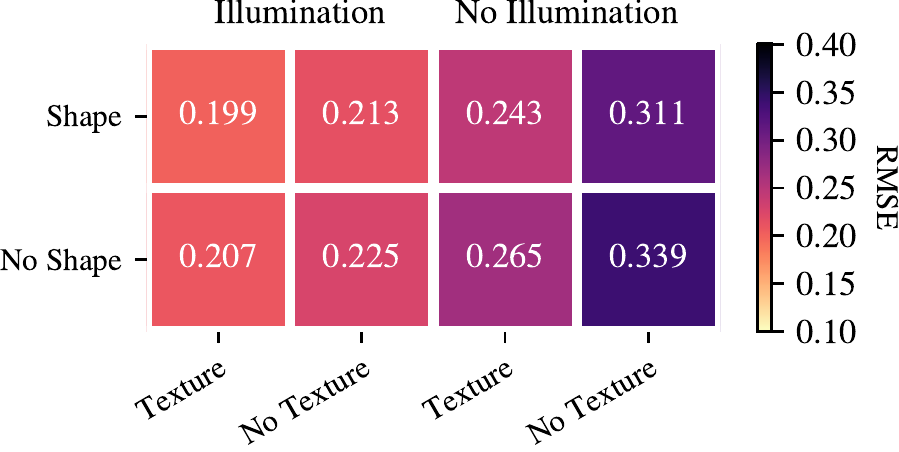}
\caption{Average RMSE on testing data collected in the real world (lower is better). Headers indicate what is randomized during data collection.}
\label{fig:ablation}
\end{figure}

During data collection in simulation, we perform randomization of background, illumination conditions, and gate appearance (shape and texture).
In this experiments, we study the effect of each of the randomized factors, except for the background which is well known to be fundamental for transfer~\cite{james2017transferring,tobin2017domain,cad2rl}.
We use as metric the Root Mean Square Error (RMSE) in prediction on our collected dataset.
As shown in Figure~\ref{fig:ablation}, illumination is the most important of the randomization factors, while gate shape randomization has the smallest effect.
Indeed, while gate appearance is similar in the real world and in simulation, the environment appearance and illumination are drastically different.
However, including more randomization is always beneficial for the robustness of the resulting policy (Figure~\ref{fig:simulation_generalization}).

\section{Discussion and Conclusion}
\label{sec:conclusions}
We have presented a new approach to autonomous, vision-based drone
racing. 
Our method uses a compact convolutional neural network to continuously predict a desired waypoint and speed directly from raw images. 
These high-level navigation directions are then executed by a classic planning and control pipeline.
As a result, the system combines the robust perceptual awareness of modern machine learning pipelines with the precision and speed of well-known control algorithms.

We investigated the capabilities of this integrated approach over three axes: precision, speed, and generalization.
Our extensive experiments, performed both in simulation and on a physical platform, show that our system is able to navigate complex race tracks, avoids the problem of drift that is inherent in systems relying on global state estimates, and can cope with highly dynamic and cluttered environments.

Our previous conference work~\cite{kaufmann2018deep} required collecting a substantial amount of training data from the track of interest. 
Here instead we propose to collect diverse simulated data via domain randomization to train our perception policy.
The resulting system can not only adapt to drastic appearance changes in simulation, but can also be deployed to a physical platform in the real world even if only trained in simulation.
Thanks to the abundance of simulated data, a perception system trained in simulation can achieve higher robustness to changes in environment characteristics (e.g.\ illumination conditions) than a system trained with real data.

It is interesting to compare the two training strategies---on real data and sim-to-real---in how they handle ambiguous situations in navigation, for instance when no gate is visible or multiple gates are in the field of view.
Our previous work~\cite{kaufmann2018deep}, which was trained on the test track, could disambiguate those cases by using cues in the environment, for instance discriminative landmarks in the background.
This can be seen as implicitly memorizing a map of the track in the network weights.
In contrast, when trained only in simulation on multiple tracks (or randomized versions of the same track), our approach can no longer use such background cues to disambiguate the flying direction and has instead to rely on a high-level map prior.
This prior, automatically inferred from the training data, describes some common characteristics of the training tracks, such as, for instance, to always turn right when no gate is visible.
Clearly, when ambiguous cases cannot be resolved with a prior of this type (e.g.\ an 8-shaped track), our sim-to-real approach would likely fail.
Possible solutions to this problem are fine-tuning with data coming from the real track, or the use of a metric prior on the track shape to make decisions in ambiguous conditions~\cite{kaufmann2018beauty}.

\cready{Due to modularity, our system can combine model-based control with learning-based perception. However, one of the main disadvantages of modularity is that errors coming from each sub-module degrade the full system performance in a cumulative way. To overcome this problem, we plan to improve each component with experience using a reinforcement learning approach. This could increase the robustness of the system and improve its performance in challenging scenarios (e.g.\ with moving obstacles).}

While our current set of experiments was conducted in the context of drone racing,
we believe that the presented approach could have broader implications for building
robust robot navigation systems that need to be able to act in a highly dynamic
world. 
Methods based on geometric mapping, localization, and planning have inherent
limitations in this setting. 
Hybrid systems that incorporate machine learning, like the one presented in this paper, can offer a compelling solution to this task, given the possibility to benefit from near-optimal solutions to different subproblems.
However, scaling our proposed approach to more general applications, such as disaster response or industrial inspection, poses several challenges.
First, due to the unknown characteristics of the path to be flown (layout, presence and type of landmarks, obstacles), the generation of a valid teacher policy would be impossible.
This could be addressed with techniques such as \emph{few-shot learning}.
Second, the target applications might require extremely high agility, for instance in the presence of sharp turns, which our autonomous system still lacks of.
This issue could be alleviated by integrating learning deeper into the control system~\cite{pan2018agile}.



\section*{acknowledgements}
This work was supported by the Intel Network on Intelligent Systems, the Swiss National Center of
Competence Research Robotics (NCCR), through the Swiss National Science Foundation, and the
SNSF-ERC starting grant.

\bibliographystyle{IEEEtran}
\bibliography{bibliography/references.bib}

\begin{IEEEbiography}[{\includegraphics[width=1in,height=1.25in,clip,keepaspectratio]{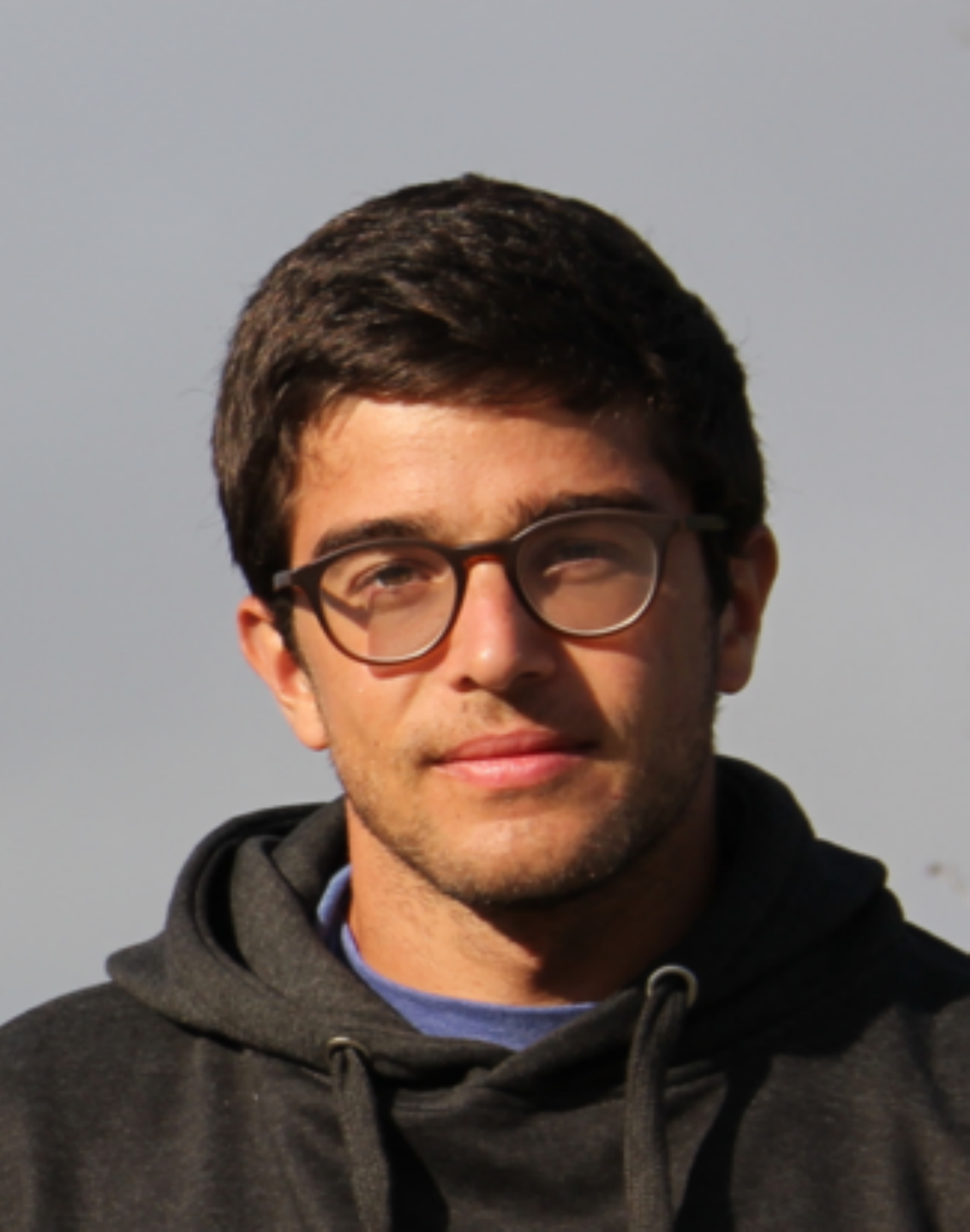}}]{Antonio Loquercio} received the MSc degree in Robotics, Systems and Control from ETH Z\"{u}rich in 2017.
He is working toward the Ph.D. degree in the Robotics and Perception Group at the University of Z\"{u}rich under the supervision of Prof. Davide Scaramuzza.
His main interests are data-driven methods for perception and control in robotics. 
He is a recipient of the ETH Medal for outstanding master thesis (2017).
\end{IEEEbiography}

\begin{IEEEbiography}[{\includegraphics[width=1in,height=1.25in,clip,keepaspectratio]{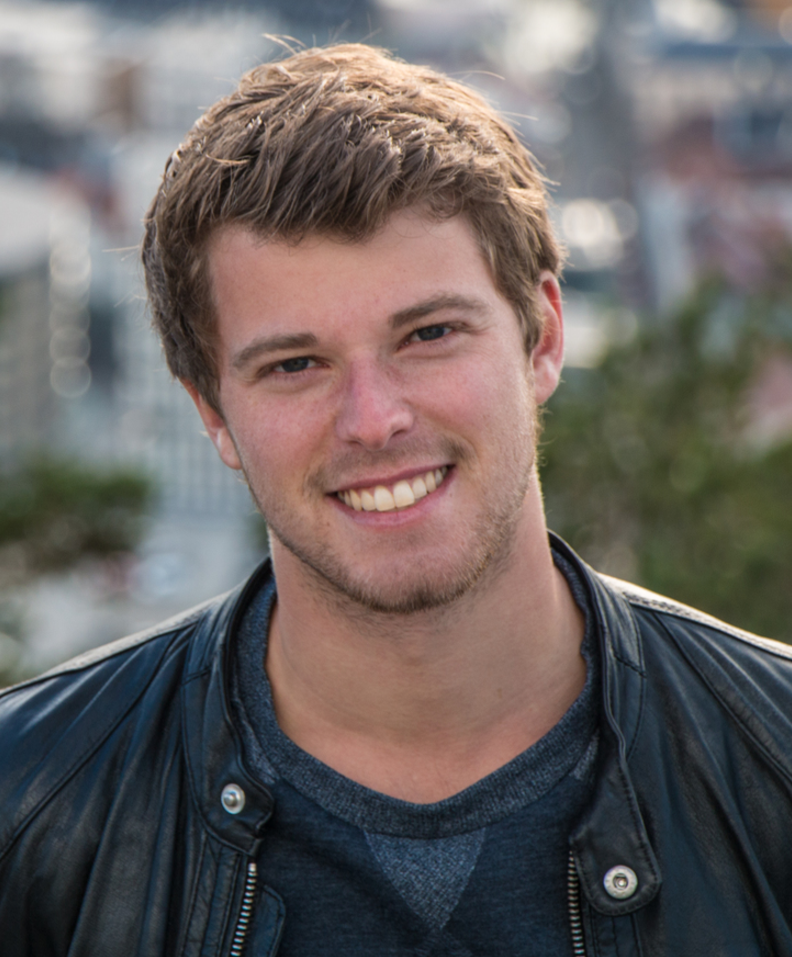}}]{Elia Kaufmann} (1992, Switzerland) obtained the M.Sc. degree in Robotics, Systems and Control at ETH Zurich, Switzerland in 2017. Previously, he received a B.Sc. degree in Mechanical Engineering (2014). Since 2017, he is pursuing a Ph.D. degree in Computer Science at the University of Zurich under the supervision of Davide Scaramuzza. He is broadly interested in the application of machine learning to improve perception and control of autonomous mobile robots.
\end{IEEEbiography}

\begin{IEEEbiography}[{\includegraphics[width=1in,height=1.25in,clip,keepaspectratio]{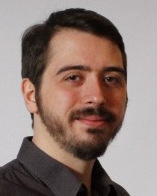}}]{Ren\'e Ranftl} is a Senior Research Scientist at the Intelligent Systems Lab at Intel in Munich, Germany. He received an M.Sc. degree and a Ph.D. degree from Graz University of Technology, Austria, in 2010 and 2015, respectively. His research interest broadly spans topics in computer vision, machine learning, and robotics. 
\end{IEEEbiography}

\begin{IEEEbiography}[{\includegraphics[width=1in,height=1.25in,clip,keepaspectratio]{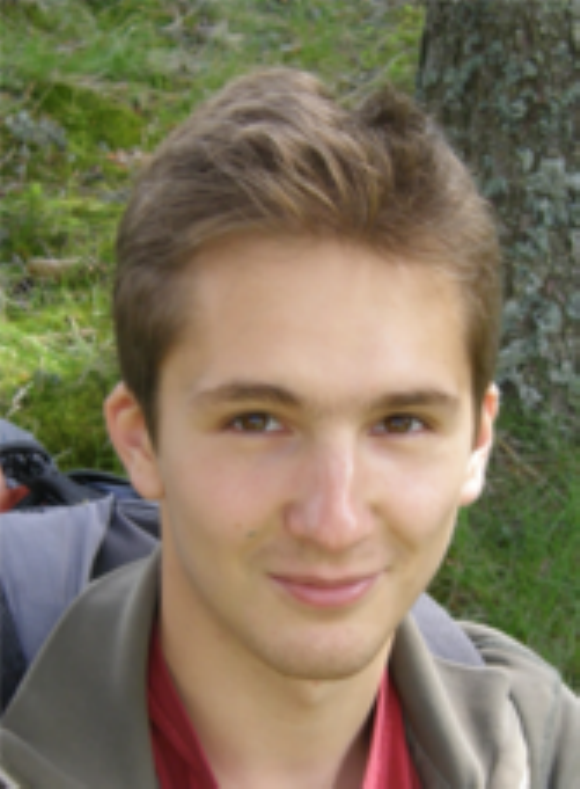}}]{Alexey Dosovitskiy} is a Research Scientist at Intel Labs. He received MSc and PhD degrees in mathematics (functional analysis) from Moscow State University in 2009 and 2012 respectively. Alexey then spent 2013-2016 as a postdoctoral researcher with Prof. Thomas Brox at the Computer Vision Group of the University of Freiburg, working on various topics in deep learning, including self-supervised learning, image generation with neural networks, motion and 3D structure estimation. In 2017 Alexey joined Intel Visual Computing Lab led by Dr. Vladlen Koltun, where he worked on applications of deep learning to sensorimotor control, including autonomous driving and robotics.

\end{IEEEbiography}

\begin{IEEEbiography}[{\includegraphics[width=1in,height=1.25in,clip,keepaspectratio]{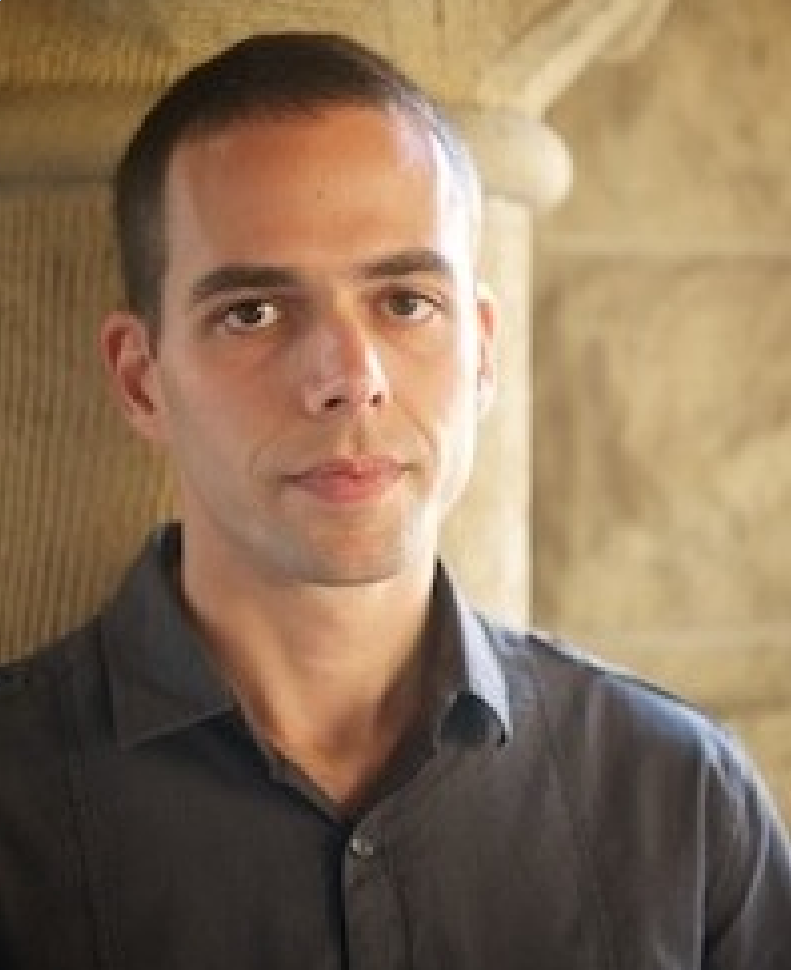}}]{Vladlen Koltun} is a Senior Principal Researcher and the director of
the Intelligent Systems Lab at Intel. His lab conducts high-impact
basic research on intelligent systems, with emphasis on computer
vision, robotics, and machine learning. He has mentored more than 50
PhD students, postdocs, research scientists, and PhD student interns,
many of whom are now successful research leaders.
\end{IEEEbiography}

\begin{IEEEbiography}[{\includegraphics[width=1in,height=1.25in,clip,keepaspectratio]{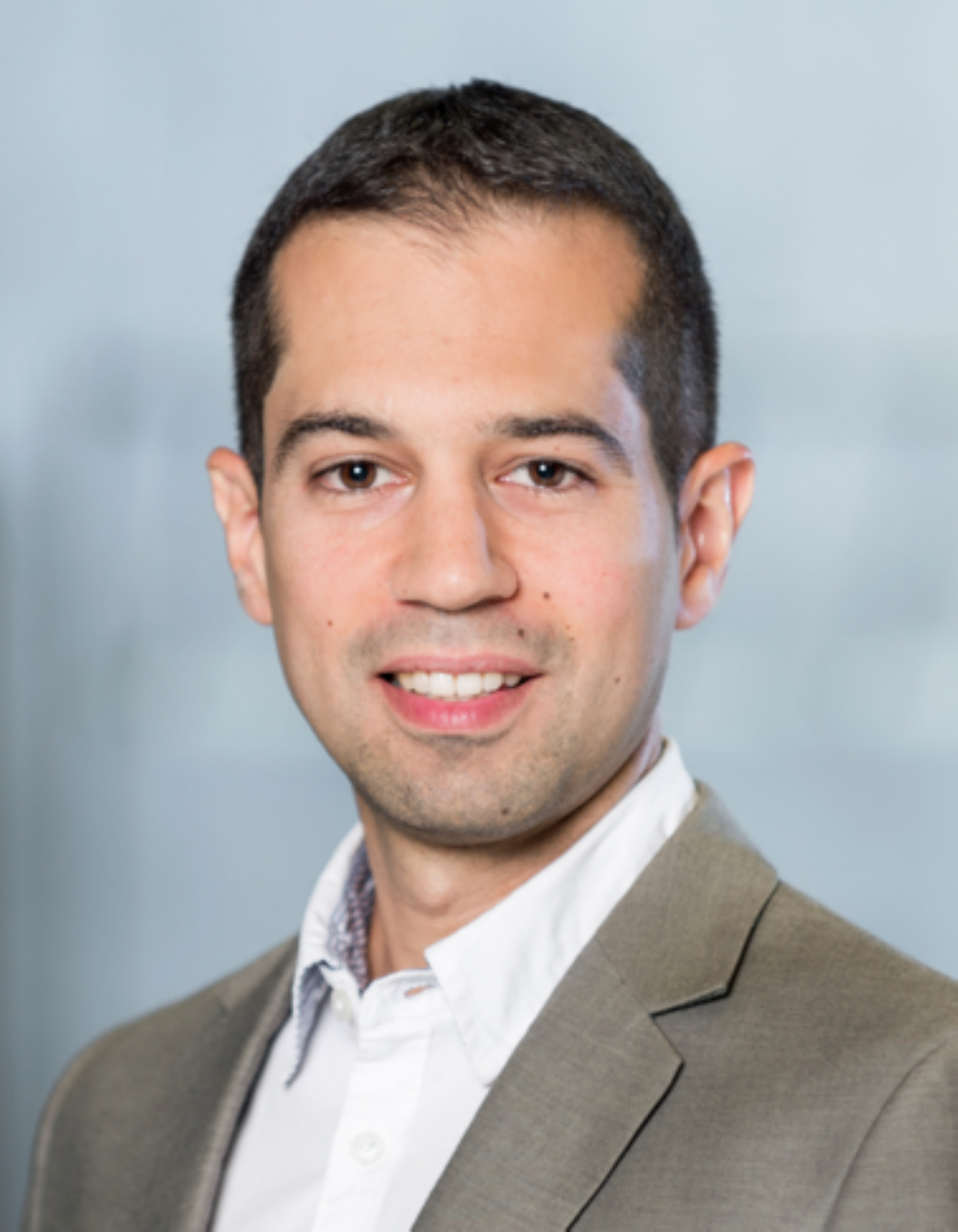}}]{Davide Scaramuzza} (1980, Italy) received the Ph.D. degree in robotics and computer vision from ETH Z\"{u}rich, Z\"{u}rich, Switzerland, in 2008, and a Postdoc at University of Pennsylvania, Philadelphia, PA, USA. He is a Professor of Robotics with University of Z\"{u}rich, where he does research at the intersection of robotics, computer vision, and neuroscience.
From 2009 to 2012, he led the European project sFly, which introduced the world's first autonomous navigation of microdrones in GPS-denied environments using visual-inertial sensors as the only sensor modality.
He coauthored the book Introduction to Autonomous Mobile Robots (MIT Press). 
Dr. Scaramuzza received an SNSF-ERC Starting Grant, the IEEE Robotics and Automation Early Career Award, and a Google Research Award for his research contributions.
\end{IEEEbiography}

\clearpage
\section*{Appendix}

\begin{figure*}[ht]
\centering
\def\colwidth{0.49\columnwidth}
\addtolength{\tabcolsep}{-4pt}
\begin{tabular}{m{\colwidth} m{\colwidth} m{\colwidth} m{\colwidth}}
\centering
  \includegraphics[width=\linewidth]{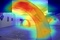} &
  \includegraphics[width=\linewidth]{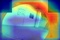} &
  \includegraphics[width=\linewidth]{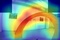} &
  \includegraphics[width=\linewidth]{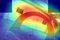} \\
\end{tabular}
\addtolength{\tabcolsep}{4pt}
\vspace{-2ex}
\captionof{figure}{Visualization of network attention using the Grad-CAM technique \cite{gradcam}. Yellow to red
areas correspond to areas of medium to high attention, while blue corresponds to areas of low attention.
It is evident that the network learns to mostly focus on gates instead of relying on the background, which explains
its capability to robustly handle dynamically moving gates. (Best viewed in color.)}
\label{fig:gradcams}
\end{figure*}
\begin{figure*}[ht]
\centering
\def\colwidth{0.49\columnwidth}
\addtolength{\tabcolsep}{-4pt}
\begin{tabular}{m{\colwidth} m{\colwidth} m{\colwidth} m{\colwidth}}
\centering
  \includegraphics[width=\linewidth]{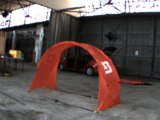} &
  \includegraphics[width=\linewidth]{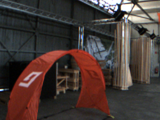} &
  \includegraphics[width=\linewidth]{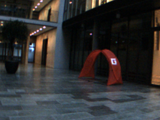} &
  \includegraphics[width=\linewidth]{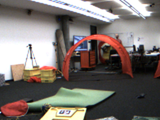} \\
\end{tabular}
\addtolength{\tabcolsep}{4pt}
\vspace{-2ex}
\captionof{figure}{Samples from dataset used in the ablation studies to quantify the importance of the randomization factors. (Best viewed in color.)}
\label{fig:ablation_dataset}
\end{figure*}

\subsection{Gamma Evaluation}
\begin{figure}
\centering
%
%
\definecolor{mycolor1}{rgb}{0.00000,0.44700,0.74100}%
\definecolor{mycolor2}{rgb}{0.85000,0.32500,0.09800}%
\definecolor{mycolor3}{rgb}{0.92900,0.69400,0.12500}%
\definecolor{mycolor4}{rgb}{0.49400,0.18400,0.55600}%
\definecolor{mycolor5}{rgb}{0.46600,0.67400,0.18800}%
\definecolor{mycolor6}{rgb}{0.30100,0.74500,0.93300}%
\definecolor{mycolor7}{rgb}{0.63500,0.07800,0.18400}%
\begin{tikzpicture}

\begin{axis}[%
width=3.0in,
height=1.2in,
at={(0.758in,0.481in)},
scale only axis,
bar shift auto,
xmin=0.5,
xmax=7.5,
xtick={1,2,3,4,5},
xticklabels={{1-20},{21-40},{41-60},{61-80},{81-100}},
xlabel style={font=\color{white!15!black}},
xlabel={Epochs trained},
ymin=0,
ymax=100,
axis background/.style={fill=white},
title style={font=\bfseries},
title={\empty},
xmajorgrids,
ymajorgrids,
legend style={legend cell align=left, align=left, draw=white!15!black, font=\scriptsize}
]
\addplot[ybar, bar width=0.05, fill=mycolor1, draw=mycolor1, area legend] table[row sep=crcr] {%
1	10.3819444444444\\
2	53.9583333333333\\
3	65.8680555555556\\
4	81.875\\
5	94.375\\
};
\addplot[forget plot, color=white!15!black] table[row sep=crcr] {%
0.5	0\\
7.5	0\\
};
\addlegendentry{$\gamma\text{ = 1e2}$}

\addplot[ybar, bar width=0.05, fill=mycolor2, draw=mycolor2, area legend] table[row sep=crcr] {%
1	42.9662698412698\\
2	96.7013888888889\\
3	98.5416666666667\\
4	99.2708333333333\\
5	96.1111111111111\\
};
\addplot[forget plot, color=white!15!black] table[row sep=crcr] {%
0.5	0\\
7.5	0\\
};
\addlegendentry{$\gamma\text{ = 1e1}$}

\addplot[ybar, bar width=0.05, fill=mycolor3, draw=mycolor3, area legend] table[row sep=crcr] {%
1	83.6309523809524\\
2	96.5972222222222\\
3	95.625\\
4	100\\
5	100\\
};
\addplot[forget plot, color=white!15!black] table[row sep=crcr] {%
0.5	0\\
7.5	0\\
};
\addlegendentry{$\gamma\text{ = 1e0}$}

\addplot[ybar, bar width=0.05, fill=mycolor4, draw=mycolor4, area legend] table[row sep=crcr] {%
1	84.9206349206349\\
2	99.0972222222222\\
3	100\\
4	100\\
5	100\\
};
\addplot[forget plot, color=white!15!black] table[row sep=crcr] {%
0.5	0\\
7.5	0\\
};
\addlegendentry{$\gamma\text{ = 1e-1}$}

\addplot[ybar, bar width=0.05, fill=mycolor5, draw=mycolor5, area legend] table[row sep=crcr] {%
1	58.4781746031746\\
2	93.0902777777778\\
3	87.1527777777778\\
4	83.2986111111111\\
5	95.625\\
};
\addplot[forget plot, color=white!15!black] table[row sep=crcr] {%
0.5	0\\
7.5	0\\
};
\addlegendentry{$\gamma\text{ = 1e-2}$}

\addplot[ybar, bar width=0.05, fill=mycolor6, draw=mycolor6, area legend] table[row sep=crcr] {%
1	67.8115079365079\\
2	87.1875\\
3	91.0416666666667\\
4	99.5486111111111\\
5	100\\
};
\addplot[forget plot, color=white!15!black] table[row sep=crcr] {%
0.5	0\\
7.5	0\\
};
\addlegendentry{$\gamma\text{ = 1e-3}$}

\addplot[ybar, bar width=0.05, fill=mycolor7, draw=mycolor7, area legend] table[row sep=crcr] {%
1	47.1795634920635\\
2	93.0902777777778\\
3	93.6805555555556\\
4	100\\
5	96.5555555555556\\
};
\addplot[forget plot, color=white!15!black] table[row sep=crcr] {%
0.5	0\\
7.5	0\\
};
\addlegendentry{$\gamma\text{ = 1e-4}$}

\end{axis}
\end{tikzpicture}%
\caption{Success rate for different values of $\gamma$.
For each $\gamma$, the network is trained up to 100 epochs. 
Performance is evaluated after each training epoch according to the performance criterion defined in~\ref{sec:sim_experiments}.
For readability reasons, performance measurements are averaged over 20 epochs. 
}
\label{fig:gamma_validation}
\end{figure}
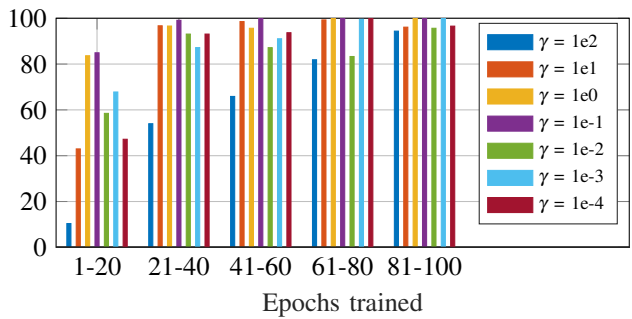

In this section, we examine the effect of the weighting factor $\gamma$ in the loss function used to train our system (Eq.~\eqref{eq:loss}).
Specifically, we selected 7 values of $\gamma$ in the range $[0.0001,100]$ equispaced in logarithmic scale.
Our network is then trained for 100 epochs on data generated from the static simulated track (Figure~\ref{img:sim_track_large}). 
After each epoch, performance is tested at a speed of $\SI{8}{\meter\per\second}$ according to the performance measure defined in~\ref{sec:sim_experiments}.
Figure~\ref{fig:gamma_validation} shows the results of this evaluation.
The model is able to complete the track for all configurations after 80 epochs.
Despite some values of $\gamma$ lead to faster learning, we see that the system performance is not too sensitive to this weighting factor.
Since $\gamma=0.1$ proves to give the best results, we use it in all our experiments.
 
\subsection{Network Architecture and Grad-CAM}

We implement the perception system using a convolutional network.
The input to the network is a $300\times200$ pixel RGB image, captured from the onboard camera at a frame rate of $\SI{30}{\hertz}$.
After normalization in the $[0,1]$ range, the input is passed through 7 convolutional layers, divided in 3 residual blocks, and a final fully connected layer that
outputs a tuple~$\lbrace \vec{x}, v \rbrace$.
$\vec{x} \in [-1,1]^2$ is a two-dimensional vector that encodes the direction to the new goal in normalized image coordinates and $v \in [0,1]$ is a normalized desired speed to approach it.

To understand why the network is robust to previously unseen changes in the environment, we visualize the network's attention using the Grad-CAM technique~\cite{gradcam} in Figure~\ref{fig:gradcams}. 
Grad-CAM visualizes which parts of an input image were important for the decisions made by the network.
It becomes evident that the network bases its decision mostly on the visual input that is
most relevant to the task at hand -- the gates -- while mostly ignoring the background.

\subsection{Additional Evaluation Dataset}

To quantify the performance of the policy trained in simulation to zero-shot generalization in real world scenarios, we collected a dataset of approximately $10$k images from the real world.
This dataset was collected from three indoor environments of different dimension and appearance.
During data collection, illumination conditions differ either for intra-day variations in natural light or for the deployment of artificial light sources.
To generate ground truth, we use the same annotation process as described in Section~\ref{sec:methodology}.
Some samples from this dataset are shown in Fig.~\ref{fig:ablation_dataset}.

%





\end{document}